\documentclass[11pt]{article}

\usepackage[preprint]{acl}

\usepackage{times}
\usepackage{latexsym}

\usepackage[T1]{fontenc}

\usepackage[utf8]{inputenc}

\usepackage{microtype}

\usepackage{inconsolata}

\usepackage{graphicx}

\usepackage{microtype}
\usepackage{graphicx}
\usepackage{subcaption}
\usepackage{booktabs} 
\usepackage{multirow}
\usepackage{makecell}

\usepackage{hyperref}

\usepackage{bm}
\usepackage{amsmath}
\usepackage{amssymb}
\usepackage{spverbatim}
\usepackage{tcolorbox}
\usepackage{dsfont}
\usepackage{float}

\usepackage{amsmath}
\usepackage{amssymb}
\usepackage{mathtools}
\usepackage{amsthm}

\usepackage[textsize=tiny]{todonotes}

\title{The Hard Decision Layer: Evidence for Committed Inference in Transformers}

\author{Ashwath {Vaithinathan Aravindan} \\
  University of Southern California\\Los Angeles, 90007, California,\\United States of America \\
  \texttt{vaithina@usc.edu} \\\And
  Mayank Kejriwal \\
  Information Sciences Institute\\4676 Admiralty Way \#1001,\\Los Angeles, 90292, California,\\United States of America \\
  \texttt{kejriwal@isi.edu} \\}

\begin{document}

\maketitle
\begin{abstract}
We investigate where and how transformer-based language models commit to predictions in multiple-choice question answering. We identify the \emph{Hard Decision Layer} (HDL), a natural architectural property where answer option rankings stabilize abruptly during inference. Empirical validation across four language models (Qwen, Llama, Granite, Mistral) and four benchmark datasets demonstrates consistent HDL emergence without learned routing policies. We also show that the HDL is invariant to fine-tuning. Our results reveal striking accuracy improvements at the HDL: up to +0.61 (Qwen on CommonsenseQA), after which performance stabilizes. Systematic ablations on label formats and problem complexity confirm the phenomenon is fundamental to model architecture. These findings offer mechanistic insights into transformer inference and suggest opportunities for efficient reasoning and model steering. All code and results required to reproduce this work are available in \url{https://github.com/Mystic-Slice/hard-decision-layer}
\end{abstract}

\begin{figure}[h!]
	\centering
	\includegraphics[width=0.47\textwidth]{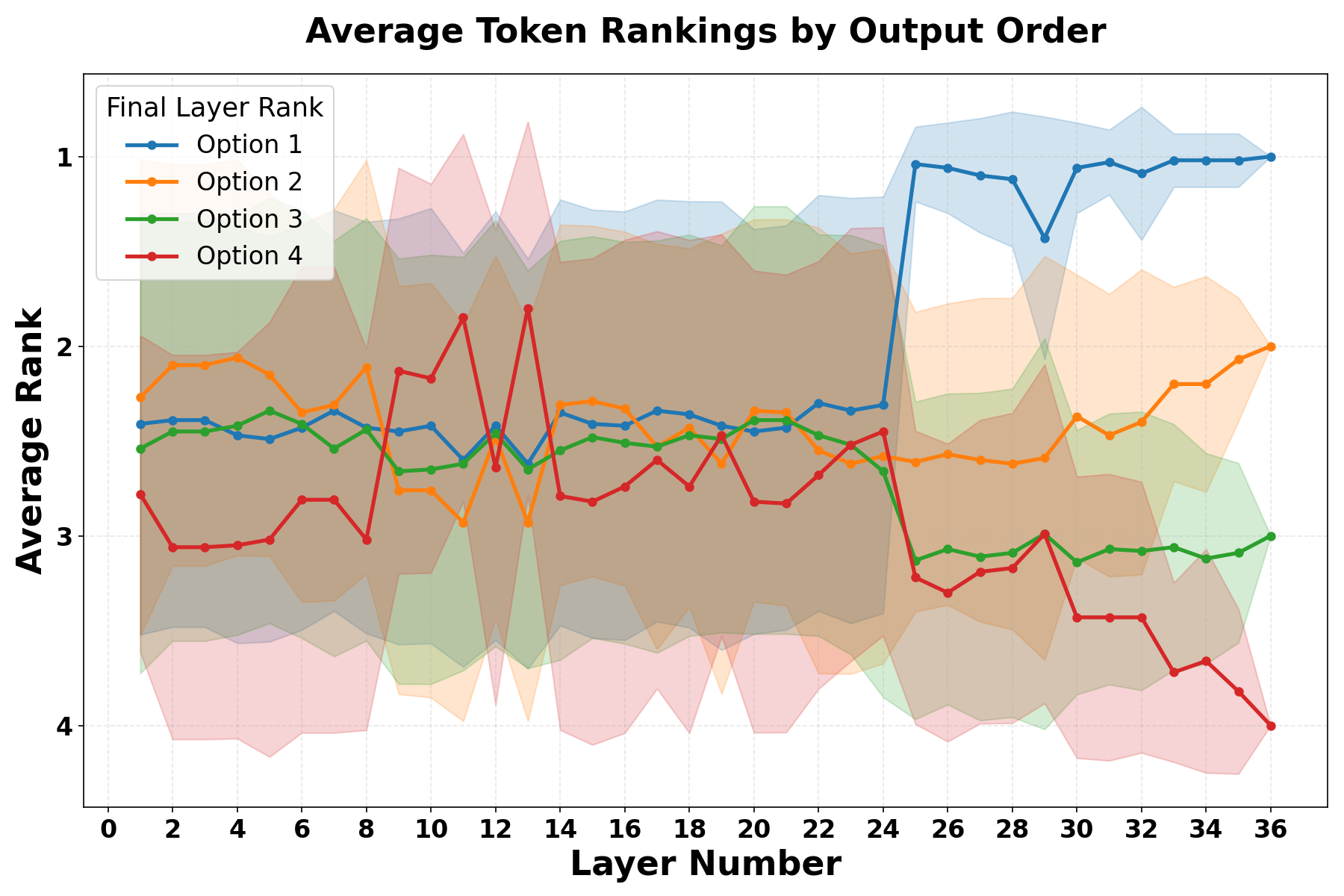}
	\caption{Average token rankings for answer option tokens across transformer layers of the Qwen model. Each line represents one of the four answer options, with average rankings computed at each layer over all questions in the QASC dataset. The faded regions surrounding each ranking line indicate the standard deviation of rankings across the dataset. Per our definition, the \textit{hard decision layer }(HDL) is identified at Layer 25. Following this layer, the average rankings are stabilized all the way until the last layer.}
	\label{fig:avg_rankings_qasc_qwen}
\end{figure}

\section{Introduction}
Understanding where and how transformer-based language models commit to their predictions is fundamental to both mechanistic interpretability~\cite{bereska2024mechanistic} and computational efficiency. Recent work shows that models allocate computational depth dynamically: information flows through the residual stream~\cite{elhage2021mathematical} and is iteratively refined by attention and feed-forward networks~\cite{geva2022transformer}, with early layers performing vocabulary filtering while later layers refine candidates~\cite{gupta2025llms}. A key insight is that model predictions stabilize mid-network~\cite{lioubashevski2024looking}, motivating research into when and how models make their final decisions.

Existing approaches to understanding layer-wise decision-making fall into two categories: dynamic methods that identify when to exit early through learned routing policies~\cite{din2023jump}, and static analyses that examine post-hoc layer pruning~\cite{fan2024not}. However, neither characterizes how models inherently organize their reasoning. What is missing is an understanding of whether there exists a natural architectural property that governs when predictions stabilize, independent of learned routing mechanisms or post-hoc modifications.

We investigate this question in the context of multiple-choice question answering (MCQA)~\cite{hendrycks2021measuring}, a controlled setting that isolates discrete reasoning from open-ended generation. We discover that a fixed \textit{Hard Decision Layer} (HDL) emerges in transformer models, beyond which the relative ranking of answer options remains stable. The HDL represents a natural, static property of how models organize their reasoning, consistent across diverse datasets and model architectures.

We make the following key contributions: (i) we identify and characterize the \textit{Hard Decision Layer (HDL)}, a natural architectural property in transformer models where answer option rankings stabilize during multiple-choice reasoning, emerging without learned routing policies. (ii) Through a comprehensive experimental study, we demonstrate that HDL is invariant to model fine-tuning (via LoRA) and robust to the number of options, confirming it reflects an intrinsic architectural property rather than a training artifact. (iii) Through systematic evaluation across four language models and four datasets, we also show that HDL prominence depends critically on option label representation, with alphabetic and arabic numeral labels producing sharper decision boundaries than roman numerals.

\section{Related Work}

Understanding how LLMs process and generate predictions is fundamental to both mechanistic interpretability and computational efficiency. Our work bridges these two areas by examining layer-wise decision-making in transformers, specifically through the lens of multiple-choice question answering (MCQA).

\subsection{Mechanistic Interpretability}

Mechanistic interpretability seeks to understand the internal computations underlying transformer predictions~\cite{bereska2024mechanistic}. Information flows through the \textit{residual stream}~\cite{elhage2021mathematical}, iteratively refined by attention and feed-forward networks (FFN) across layers. Projection-based techniques like the \textit{logit lens}~\cite{nostalgebraist2020logitlens} and \textit{tuned lens}~\cite{belrose2023eliciting} enable ``peeking'' at intermediate predictions. Complementary approaches such as activation patching and causal tracing~\cite{meng2022locating} identify causal components and factual associations~\cite{hewitt2019structural, haviv2023understanding, dar2023analyzing}.

In transformers, feed-forward layers function as key-value memories~\cite{geva2022transformer, geva2020transformer}, promoting concepts to affect outputs. Early layers perform vocabulary filtering while later layers refine candidates~\cite{gupta2025llms}. Notably, models allocate computational depth dynamically: tokens at the start of factual sequences use more layers for decision-making, while those naturally following use fewer layers, suggesting that models self-organize reasoning based on task context.

\subsection{Token Saturation Phenomenon}

Model predictions stabilize mid-network, motivating early exit research. \cite{din2023jump} use linear transformations to shortcut intermediate computation, achieving ~13.8\% layer savings on GPT-2, though this requires dynamic decisions at each layer. More structured approaches show that token ranking saturation follows a predictable order across architectures~\cite{lioubashevski2024looking}, suggesting principled decision progression and natural boundaries. \cite{fan2024not} further demonstrate that not all layers are equally essential, achieving layer pruning ratios up to 17.8\% in large models like Llama2 and OPT.

\subsection{Multiple-Choice Question Answering and Text Classification using LLMs}
MCQA benchmarks have become standard for evaluating language models~\cite{hendrycks2021measuring}, offering a balanced evaluation framework that avoids open-ended generation complexities while probing model capabilities. The use of large language models (LLMs) as classifiers through in-context learning (ICL) was established by \citet{brown2020language}, who demonstrated that GPT-3 (175B parameters) could perform sentiment analysis, natural language inference, and topic classification by conditioning on a few labeled demonstrations in the prompt and reading the next predicted token as the class label, without any gradient updates. This one-token prediction paradigm, however, was shown by \citet{zhao2021calibrate} to suffer from systematic biases: the model's probability distribution over candidate label tokens is skewed by majority-label bias (favoring whichever class appears most in the demonstrations), recency bias (favoring the label of the last demonstration), and common-token bias (favoring label words that are more frequent in pretraining). Their proposed fix, \emph{contextual calibration}, estimates these biases by feeding a content-free input (e.g., ``N/A'') and learning an affine transformation on the label logits, recovering up to 30 percentage points of accuracy. 

Our central research question is: How do modern LLMs make their answer choice in MCQA tasks? This study also extends to classification tasks using LLMs. We address this question by identifying a fixed \textit{Hard Decision Layer} (HDL)---the layer beyond which answer option rankings remain stable. Unlike prior work that relies on dynamic early exit policies or post-hoc layer pruning, the HDL emerges as a natural, static property of transformer inference without requiring learned routing mechanisms. We demonstrate that this phenomenon generalizes consistently across model architectures and evaluation datasets. These findings reveal that transformers inherently organize their inference according to architectural and task characteristics, offering mechanistic insights into how depth is utilized during reasoning.
\section{Methodology}
\subsection{Preliminary}
\subsubsection{Logit Lens}
\label{sec:logit_lens}
Logit Lens is a technique where intermediate layer representations from a transformer model are projected into the vocabulary space, allowing us to observe what output the model "predicts" at each layer. This technique enables us to track how the model's predictions evolve across layers and identify key decision points in the reasoning process. 

Let $\mathcal{M}$ denote a transformer-based large language model with $L$ layers, hidden dimension $d$, and vocabulary $\mathcal{V}$ of size $|\mathcal{V}| = V$. Given an input sequence of $T$ tokens $\bm{x} = (x_1, x_2, \dots, x_T)$ with $x_t \in \mathcal{V}$, the embedding layer produces an initial representation $\bm{h}^{(0)} \in \mathbb{R}^{T \times d}$. Each transformer block $\ell \in \{1, \dots, L\}$ updates the residual stream as
\begin{equation}
    \bm{h}^{(\ell)} = f_\ell\!\bigl(\bm{h}^{(\ell-1)}\bigr),
\end{equation}
where $f_\ell$ encapsulates multi-head self-attention followed by a position-wise feedforward network, with residual connections and layer normalization applied according to the model's specific architecture. We capture $\bm{h}^{(\ell)}$ for every $\ell$, storing the output tensor during a single forward pass with gradient computation disabled.

To interpret the intermediate representations, we project each layer's residual stream into vocabulary space using the model's unembedding head. For the token at position $T$, we first extract the hidden-state vector $\bm{h}_t^{(\ell)} \in \mathbb{R}^d$ from layer $\ell$. Because the unembedding matrix $\bm{W}_U \in \mathbb{R}^{V \times d}$ is trained to operate on the output of the final layer normalization, we apply the model's final-layer normalization function $\text{LN}_\text{final}(\cdot)$ before projection:
\begin{equation}
    \hat{\bm{h}}_t^{(\ell)} = \text{LN}_\text{final}\!\bigl(\bm{h}_t^{(\ell)}\bigr).
\end{equation}
The logit vector over the vocabulary at layer $\ell$ and position $t$ is then
\begin{equation}
    \bm{z}^{(\ell)}_t = \bm{W}_U\, \hat{\bm{h}}_t^{(\ell)} + \bm{b}_U,
\end{equation}
where $\bm{b}_U \in \mathbb{R}^V$ is the bias term if present. For each layer, we record the logits for the answer option tokens $S = \{A, B, C, D, ...\}$, yielding the pairs $\bigl\{\!\bigl(v,\, z^{(\ell)}_{t,v}\bigr) : v \in \mathcal{S}\bigr\}$.

\subsubsection{Logit Lens Extraction and Analysis}
To investigate how models arrive at answers across layers, we apply  Logit Lens to extract layer-wise logits, and thereby, token rankings for the answer options in response to the prompt\footnote{We reproduce a standard multiple-choice prompt template in Figure~\ref{fig:prompt_example} in Appendix~\ref{sec:prompt-models-datasets}, demonstrating the structured format used in our experiments.}. For each layer, we record the logits for the answer option tokens and rank them to determine the model's preference at that depth. These layer-wise rankings allow us to compute accuracy at each layer and track how the model's answer evolves throughout the network, providing mechanistic insight into the internal reasoning process.

We now formalize the ranking and accuracy computations used in our analysis. For a question $s$ with four answer options represented by vocabulary tokens $v_1, v_2, v_3, v_4 \in \mathcal{V}$, we extract logits at each layer $\ell$ using the Logit Lens procedure.

At layer $\ell$ and position $T$ (the position of the answer token in the sequence), let $\bm{z}^{(\ell)}_T$ denote the full logit vector over vocabulary, as defined in Equation~(3). We extract the logits corresponding to the four answer option tokens as
\begin{equation}
\mathbf{z}_\ell = \bigl(z^{(\ell)}_{T, v_1}, z^{(\ell)}_{T, v_2}, z^{(\ell)}_{T, v_3}, z^{(\ell)}_{T, v_4}\bigr),
\end{equation}
where $z^{(\ell)}_{T, v_i}$ is the $v_i$-th component of $\bm{z}^{(\ell)}_T$. We compute the ranking by sorting these logits in descending order:
\begin{equation}
\text{rank}_\ell(v_i) = \text{argsort}_{\text{descending}}(\mathbf{z}_\ell)(v_i),
\end{equation}
where $\text{rank}_\ell(v_i) \in \{1, 2, 3, 4\}$ represents the position of option token $v_i$ in the sorted logits at layer $\ell$, with rank 1 indicating the highest logit.

To aggregate layer-wise rankings across the dataset, we group tokens by their rank at the final output layer. Let $\mathcal{D}$ denote the set of all questions, and for each question $s \in \mathcal{D}$, let $\text{output\_rank}(s, v_i)$ denote the output option rank (position in the final-layer ranking) of token $v_i$. Let $I_k(s, v_i) := \mathds{1}[\text{output\_rank}(s, v_i) = k]$, where $\mathds{1}[\cdot]$ is the indicator function, be the indicator that token $v_i$ achieves rank $k$ in question $s$. The average rank of tokens with output option rank $k$ at layer $\ell$ is computed as:
\begin{equation}\label{eqn:avg_rank}
\overline{\text{rank}}_\ell(k) = \frac{1}{|\mathcal{D}|} \sum_{s \in \mathcal{D}} \sum_{i=1}^{4} \text{rank}_\ell(v_i) \cdot I_k(s, v_i),
\end{equation}
with standard deviation $\sigma_\ell(k)$ computed across questions to capture variability.

Finally, layer-wise accuracy measures the fraction of questions for which the correct answer token achieves the highest ranking at layer $\ell$. Let $v_{\text{correct}}(s)$ denote the correct answer token for question $s$. The accuracy at layer $\ell$ is:
\begin{equation}
\text{Acc}_\ell = \frac{1}{|\mathcal{D}|} \sum_{s \in \mathcal{D}} \mathds{1}[\text{rank}_\ell(v_{\text{correct}}(s)) = 1],
\end{equation}
where $\mathds{1}[\cdot]$ is the indicator function.

\subsubsection{Hard-Decision Layer}
We identify a unique layer - called the Hard-Decision Layer (HDL) - in the modern LLMs tested, which causes the steepest drop in the average ranking of the answer option that eventually ends up as the predicted answer by the model. 
Let $\text{output\_rank}(s, v_i) = 1$ identify the final predicted answer option for question $s$. HDL is formally defined as the layer where the largest decrease in average rank occurs:
\begin{equation}
\ell_{\text{HDL}} = \arg\max_{\ell \in \{1, \ldots, L\}} \left[ \overline{\text{rank}}_\ell(1) - \overline{\text{rank}}_{\ell-1}(1) \right],
\end{equation}
where $\overline{\text{rank}}_\ell(1)$ is the average rank of the predicted answer token (output rank 1) at layer $\ell$, as defined in Equation~\ref{eqn:avg_rank}. This layer represents the critical point in the network where the model commits to its final answer, with the steepest improvement in answer ranking indicating the strongest refinement in the decision process. 

\subsection{Experimental Setup}
\subsubsection{Models}
We evaluate four instruction-tuned language models: Mistral-7B-Instruct-v0.3~\cite{Jiang2023Mistral7}, Llama-3.1-8B-Instruct~\cite{grattafiori2024llama}, IBM Granite-3.3-2B-Instruct\footnote{\url{https://huggingface.co/ibm-granite/granite-3.3-2b-instruct}}, and Qwen3-4B-Instruct-2507~\cite{yang2025qwen3}. These models range from 2B to 8B parameters and represent diverse architectures and training methodologies. More information about the models used is provided in Appendix~\ref{sec:prompt-models-datasets} (Table~\ref{tab:models}).

\subsubsection{Datasets}
We evaluate model behavior across four multiple-choice question-answering datasets: CommonsenseQA~\cite{talmor-etal-2019-commonsenseqa}, a benchmark for evaluating commonsense reasoning; QASC~\cite{allenai:qasc}, which requires compositional reasoning and multi-step inference; MMLU-Pro~\cite{wang2024mmlu}, an enhanced MMLU benchmark covering STEM, social sciences, and humanities; and SuperGPQA~\cite{du2025supergpqa}, a graduate-level benchmark covering advanced concepts in physics, chemistry, and biology. For each model and dataset, we sample 100 random questions and normalize them to contain exactly four options per question. When a question contains more than four options, we randomly select four while ensuring the correct answer remains in the pool. A detailed summary of the datasets is provided in Appendix~\ref{sec:prompt-models-datasets} (Table~\ref{tab:datasets}).

\subsubsection{Fine-tuning with Low-Rank Adaptation}
To understand whether HDL persists after fine-tuning, we fine-tune two models (Qwen and Llama) using low-rank adaptation (LoRA)~\cite{hu2022lora} for 1 epoch on a separate training set consisting of $8000$ \text{to} $10000$ samples for each evaluated dataset. Given base parameters $\bm{W}_0$ and low-rank matrices $\bm{A} \in \mathbb{R}^{d \times r}$, $\bm{B} \in \mathbb{R}^{r \times d}$ with rank $r \ll d$, a model with the LoRA adapter behaves identically to the model with $\bm{W} = \bm{W}_0 + \bm{A}\bm{B}$ for each adapted weight matrix. Details regarding the hyperparameters used for model training are provided in Appendix~\ref{sec:hyperparams}.

\subsubsection{Task Variants} 
To examine whether other aspects of the multiple-choice format affect the HDL, we conduct experiments varying both the option labeling scheme and the number of available choices. We test three labeling formats on the QASC dataset for both Qwen and Llama models: alphabetic labels (A/B/C/D), arabic numerals (1/2/3/4), and roman numerals (i/ii/iii/iv). It was ensured that, for the models tested, the new option labels are still tokenized to 1 token. We additionally vary the number of options from three to five choices (using a subset or extension of the original options) to understand how the HDL manifests across different problem complexities. For each configuration, we extract layer-wise logits and compute both option token rankings and layer-wise accuracy. This allows us to isolate the effects of notational convention and problem structure from the underlying model reasoning and assess whether the HDL is robust across different task variants.

\subsubsection{Open-ended Generation} 
To investigate whether HDL as a detectable phenomenon extends beyond multiple-choice reasoning to general sequence generation, we conduct preliminary experiments on open-ended question answering using the GSM8K~\cite{cobbe2021gsm8k} dataset of mathematical word problems. Rather than tracking option token rankings, we extract residual states during generation of the $n$-th output token at various generation steps ($n = 10, 50, 100, 200$), allowing us to observe layer-wise dynamics across different points in the generation process. For each step, we capture the top-10 most likely output tokens at each layer by applying Logit Lens to the intermediate representations. This analysis allows us to assess whether intermediate layers exhibit stabilization patterns similar to those observed in multiple-choice settings and to examine how the rank of the generated token evolves across layers in a free-form generation context.

\section{Results}
\textbf{Finding 1: HDL exists in all models and is invariant across MCQA datasets.}

\begin{table*}
\centering
\small
\resizebox{\textwidth}{!}{
\begin{tabular}{llcccccc}
\toprule
\textbf{Model} & \textbf{Dataset} & \makecell{\textbf{HDL Layer} \\ \textbf{(Predicted)}} & \makecell{\textbf{Accuracy} \\ \textbf{(Pre-HDL)}} & \makecell{\textbf{Accuracy} \\ \textbf{(Post-HDL)}} & \makecell{\textbf{Accuracy} \\ \textbf{(Final)}} & \makecell{\textbf{Option-1 Rank} \\ \textbf{(Pre-HDL)}} & \makecell{\textbf{Option-1 Rank} \\ \textbf{(Post-HDL)}} \\
\midrule
\multirow{4}{*}{Qwen} & QASC & 25 & 0.31 & 0.70 (+0.39) & 0.72 (+0.02) & 2.31 $\pm$ 1.09 & 1.04 $\pm$ 0.20 \\
 & MMLU-Pro & 25 & 0.23 & 0.58 (+0.35) & 0.61 (+0.03) & 2.52 $\pm$ 1.17 & 1.09 $\pm$ 0.35 \\
 & CommonsenseQA & 25 & 0.26 & 0.87 (+0.61) & 0.86 (-0.01) & 2.42 $\pm$ 1.07 & 1.01 $\pm$ 0.10 \\
 & SuperGPQA & 25 & 0.24 & 0.40 (+0.16) & 0.41 (+0.01) & 2.22 $\pm$ 1.11 & 1.05 $\pm$ 0.26 \\
\midrule
\multirow{4}{*}{Llama} & QASC & 18 & 0.28 & 0.59 (+0.31) & 0.74 (+0.15) & 2.35 $\pm$ 1.01 & 1.42 $\pm$ 0.62 \\
 & MMLU-Pro & 18 & 0.25 & 0.45 (+0.20) & 0.59 (+0.14) & 2.72 $\pm$ 1.11 & 1.97 $\pm$ 1.05 \\
 & CommonsenseQA & 18 & 0.24 & 0.63 (+0.39) & 0.80 (+0.17) & 2.54 $\pm$ 1.08 & 1.52 $\pm$ 0.81 \\
 & SuperGPQA & 23 & 0.29 & 0.41 (+0.12) & 0.32 (-0.09) & 2.11 $\pm$ 0.96 & 1.58 $\pm$ 0.89 \\
\midrule
\multirow{4}{*}{Granite} & QASC & 32 & 0.27 & 0.51 (+0.24) & 0.67 (+0.16) & 2.41 $\pm$ 1.21 & 1.56 $\pm$ 0.88 \\
 & MMLU-Pro & 32 & 0.32 & 0.45 (+0.13) & 0.47 (+0.02) & 2.54 $\pm$ 1.12 & 1.90 $\pm$ 1.01 \\
 & CommonsenseQA & 32 & 0.30 & 0.56 (+0.26) & 0.72 (+0.16) & 2.45 $\pm$ 1.20 & 1.73 $\pm$ 1.01 \\
 & SuperGPQA & 33 & 0.27 & 0.36 (+0.09) & 0.41 (+0.05) & 2.12 $\pm$ 1.18 & 1.74 $\pm$ 1.05 \\
\midrule
\multirow{4}{*}{Mistral} & QASC & 20 & 0.38 & 0.56 (+0.18) & 0.62 (+0.06) & 1.86 $\pm$ 0.96 & 1.32 $\pm$ 0.72 \\
 & MMLU-Pro & 20 & 0.30 & 0.46 (+0.16) & 0.49 (+0.03) & 2.09 $\pm$ 1.03 & 1.35 $\pm$ 0.77 \\
 & CommonsenseQA & 20 & 0.45 & 0.68 (+0.23) & 0.76 (+0.08) & 1.97 $\pm$ 1.08 & 1.36 $\pm$ 0.66 \\
 & SuperGPQA & 20 & 0.27 & 0.33 (+0.06) & 0.37 (+0.04) & 2.08 $\pm$ 0.89 & 1.50 $\pm$ 0.69 \\
\bottomrule
\end{tabular}
}
\caption{Descriptive HDL summary across models and datasets. Accuracy (Pre-HDL), Accuracy (Post-HDL), and Accuracy (Final) are mean accuracies of the logit-lens projection at layers before the predicted HDL, at and after the HDL, and at the final layer, respectively. The HDL Layer (Predicted) is predicted as the layer that causes the largest drop in Option 1's average rank. Parenthetical values in the accuracy columns indicate absolute increases: Accuracy (Post-HDL) shows the increase over Accuracy (Pre-HDL), and Accuracy (Final) shows the increase over Accuracy (Post-HDL). A small final-post gap indicates that post-HDL behavior already matches the model's final output. Option-1 Rank (Pre-HDL) and Option-1 Rank (Post-HDL) show the mean and standard deviation of the rank of Option 1 at the layer before HDL and at the HDL layer, respectively. Lower rank values indicate greater concentration on the correct answer, demonstrating the rank collapse that shows the model committing to an answer.}
\label{tbl:hdl_summary}

\end{table*}

As shown in Table~\ref{tbl:hdl_summary}, all models have a fixed HDL for the MCQA datasets tested. Figure~\ref{fig:hdl_vs_dataset_qwen} and Figure~\ref{fig:hdl_vs_model_qasc} in Appendix~\ref{sec:supp-finding-1} show the generalization of this behaviour across datasets and models respectively. Some models position their HDL early in the network (e.g., Llama at layer 18 of 32 total layers, or 56.25\% depth), while others place it much later (e.g., Granite at layer 32 of 40 total layers, or 80\% depth). Table~\ref{table:hdl_statistics} summarizes the model parameter counts and the fraction of layers up to the HDL, representing where most computation occurs. A trend is observed where parameter count appears inversely related to the fraction of layers until the HDL. This suggests that higher-capacity models accomplish answer selection in earlier layers. 

\begin{table}
	\centering
	\caption{Depth of Hard Decision Layer (as percentage of layers) across models}
	\label{table:hdl_statistics}
	\begin{tabular}{lcc}
		\toprule
		\textbf{Model} & \textbf{Parameters} & \makecell{\textbf{Depth of HDL} \\ \textbf{(as \% of layers)}} \\
		\midrule
		Llama & 8B & 56.25\% (18/32) \\
    Mistral & 7B & 62.50\% (20/32)\\
		Qwen & 4B & 69.44\% (25/36)\\
		Granite & 2B & 80.00\% (32/40)\\
		\bottomrule
	\end{tabular}
\end{table} 

Table~\ref{tbl:hdl_summary} also demonstrates a sharp jump in the accuracy of intermediate layer outputs at the HDL (also illustrated in Figures~\ref{fig:hdl_qwen_accuracy} and~\ref{fig:hdl_llama_accuracy} in Appendix~\ref{sec:supp-finding-2}). For some models, such as Qwen, the accuracy at the HDL nearly matches that of the final layer. The accuracy spike at HDL is much larger than the accuracy improvement from HDL till the final output layer. Remarkably, the accuracy of layers before the HDL hovers around 25\%, which equals random chance for four-option multiple-choice questions. This suggests the model essentially selects its answer, for most questions, at the HDL. 

Citing a specific example of one dataset and model, Figure~\ref{fig:avg_rankings_qasc_qwen} shows the aggregated layer-wise rankings of the answer option tokens by the Qwen model across the entire QASC dataset. In the plot, Options 1, 2, 3, and 4 represent the options ranked 1st, 2nd, 3rd, and 4th respectively in the model's final output layer. For most questions, the model initially exhibits frequent fluctuations in the ordering of answer tokens. However, at the HDL (layer 25 for the Qwen model), the model produces a decisive ranking that remains largely stable through the remaining layers. On average, option tokens maintain their ranking from the HDL onward, suggesting that the model has completed most of its reasoning by this point. Notably, the top answer token shows greater stability than other option choices.
As shown in Figure~\ref{fig:hdl_vs_dataset_qwen}, the HDL position for a given model remains consistent across different datasets. This suggests the HDL is unlikely to be a memory storage layer, in which case we would expect its position to vary with different datasets. Rather, the HDL likely plays a role specific to multiple-choice question answering. The results also show a sharp drop in the mean and standard deviation of Option 1's rank at the HDL which shows the model committing to a specific answer.

\textbf{Finding 2: HDL is invariant both to finetuning and to the number of choice options.}
The Qwen and Llama models were finetuned on a separate training set for each of the datasets and evaluated again. The HDL position for these models remains stable even after finetuning. While the models demonstrate clear accuracy improvements (as expected) following finetuning, the HDL itself does not shift across base and finetuned variants, as confirmed quantitatively in Table~\ref{tab:hdl_finetuning_summary}, and visualized in Figures~\ref{fig:hdl_qwen_finetuning} and~\ref{fig:hdl_llama_finetuning} (Appendix~\ref{sec:supp-finding-2}). 

\begin{table*}[h!]
\centering
\small
\begin{tabular}{lllrrcc}
\toprule
\textbf{Model} & \textbf{Dataset} & \textbf{Variant} & \makecell{\textbf{HDL Layer} \\ \textbf{(Predicted)}} & \makecell{\textbf{Accuracy} \\ \textbf{(Pre-HDL)}} & \makecell{\textbf{Accuracy} \\ \textbf{(Post-HDL)}} & \makecell{\textbf{Accuracy} \\ \textbf{(Final)}} \\
\midrule
\multirow{2}{*}{Llama} & \multirow{2}{*}{CommonsenseQA} & Base & 18 & 0.24 & 0.63 (+0.39) & 0.80 (+0.17) \\
 & & Finetuned & 18 & 0.25 & 0.60 (+0.35) & 0.86 (+0.26) \\
\midrule
\multirow{2}{*}{Llama} & \multirow{2}{*}{MMLU-Pro} & Base & 18 & 0.25 & 0.45 (+0.20) & 0.59 (+0.14) \\
 & & Finetuned & 18 & 0.21 & 0.40 (+0.19) & 0.65 (+0.25) \\
\midrule
\multirow{2}{*}{Llama} & \multirow{2}{*}{QASC} & Base & 18 & 0.28 & 0.59 (+0.31) & 0.74 (+0.15) \\
 & & Finetuned & 18 & 0.29 & 0.61 (+0.32) & 0.77 (+0.16) \\
\midrule
\multirow{2}{*}{Llama} & \multirow{2}{*}{SuperGPQA} & Base & 23 & 0.29 & 0.41 (+0.12) & 0.32 (-0.09) \\
 & & Finetuned & 18 & 0.23 & 0.31 (+0.08) & 0.40 (+0.09) \\
\midrule
\multirow{2}{*}{Qwen} & \multirow{2}{*}{CommonsenseQA} & Base & 25 & 0.26 & 0.87 (+0.61) & 0.86 (-0.01) \\
 & & Finetuned & 25 & 0.27 & 0.86 (+0.59) & 0.86 (+0.00) \\
\midrule
\multirow{2}{*}{Qwen} & \multirow{2}{*}{MMLU-Pro} & Base & 25 & 0.23 & 0.58 (+0.35) & 0.61 (+0.03) \\
 & & Finetuned & 25 & 0.22 & 0.65 (+0.43) & 0.69 (+0.04) \\
\midrule
\multirow{2}{*}{Qwen} & \multirow{2}{*}{QASC} & Base & 25 & 0.31 & 0.70 (+0.39) & 0.72 (+0.02) \\
 & & Finetuned & 25 & 0.30 & 0.79 (+0.49) & 0.80 (+0.01) \\
\midrule
\multirow{2}{*}{Qwen} & \multirow{2}{*}{SuperGPQA} & Base & 25 & 0.24 & 0.40 (+0.16) & 0.41 (+0.01) \\
 & & Finetuned & 25 & 0.23 & 0.45 (+0.22) & 0.42 (-0.03) \\
\bottomrule
\end{tabular}
\caption{Descriptive HDL summary for base vs fine-tuned models. Each (model, dataset) pair appears on two rows: base and finetuned variants. HDL Layer (Predicted), Accuracy (Pre-HDL), Accuracy (Post-HDL), and Accuracy (Final) are mean logit-lens accuracies at layers before the predicted HDL, at and after the HDL, and at the final layer, respectively. Parenthetical values in the accuracy columns indicate absolute increases: Accuracy (Post-HDL) shows the increase over Accuracy (Pre-HDL), and Accuracy (Final) shows the increase over Accuracy (Post-HDL). Comparing base and finetuned variants within a (model, dataset) pair reveals whether fine-tuning shifted the HDL position or only its accuracy levels.}
\label{tab:hdl_finetuning_summary}
\end{table*}

Similarly, when examining whether the number of options in multiple-choice questions affects the HDL position by comparing three-option and five-option variants on the QASC dataset for both Qwen and Llama models, we found the HDL position to remain invariant across different option counts (Figure~\ref{fig:number_of_options_aggregate} in Appendix~\ref{sec:supp-finding-3}). While the pre-HDL accuracy differs according to random chance (33\% for three options, 20\% for five options), the HDL location and the accuracy patterns (Figure~\ref{fig:number_of_options_accuracy} in Appendix~\ref{sec:supp-finding-3}) do not significantly differ across these conditions - quantified in Table~\ref{tab:hdl_num_option_variants}.

\begin{table*}[h]
\centering
\small
\begin{tabular}{cccccc}
\toprule
\textbf{Model} & \textbf{Number of Options} & \makecell{\textbf{HDL Layer} \\ \textbf{(Predicted)}} & \makecell{\textbf{Accuracy} \\ \textbf{(Pre-HDL)}} & \makecell{\textbf{Accuracy} \\ \textbf{(Post-HDL)}} & \makecell{\textbf{Accuracy} \\ \textbf{(Final)}} \\
\midrule
Qwen & 3 & 25 & 0.32 & 0.78 (+0.46) & 0.77 ($-$0.01) \\
Qwen & 4 & 25 & 0.31 & 0.70 (+0.39) & 0.72 (+0.02) \\
Qwen & 5 & 25 & 0.22 & 0.64 (+0.42) & 0.64 (+0.00) \\
\hline
Llama & 3 & 18 & 0.34 & 0.62 (+0.28) & 0.78 (+0.16) \\
Llama & 4 & 18 & 0.28 & 0.59 (+0.31) & 0.74 (+0.15) \\
Llama & 5 & 18 & 0.24 & 0.55 (+0.31) & 0.70 (+0.15) \\
\bottomrule
\end{tabular}
\caption{Descriptive HDL summary across QASC option-count variants. All variants use alphabetic labels (A/B/C/D) on the same QASC questions; only the Number of Options differs. The default QASC cohort uses 4 options. Each row shows the HDL Layer (Predicted) and accuracies at the layer just before HDL (Accuracy (Pre-HDL)), at the HDL layer (Accuracy (Post-HDL)), and at the final layer (Accuracy (Final)) for one (model, number of options) combination; parenthetical values on the accuracy columns are absolute increases over the previous column. Results are restricted to Qwen and Llama because the 3- and 5-option residual-trace pickles are only available for those models.}
\label{tab:hdl_num_option_variants}
\end{table*}

\begin{figure*}[h!]
\centering
\begin{subfigure}{0.48\textwidth}
\centering
\includegraphics[width=\textwidth]{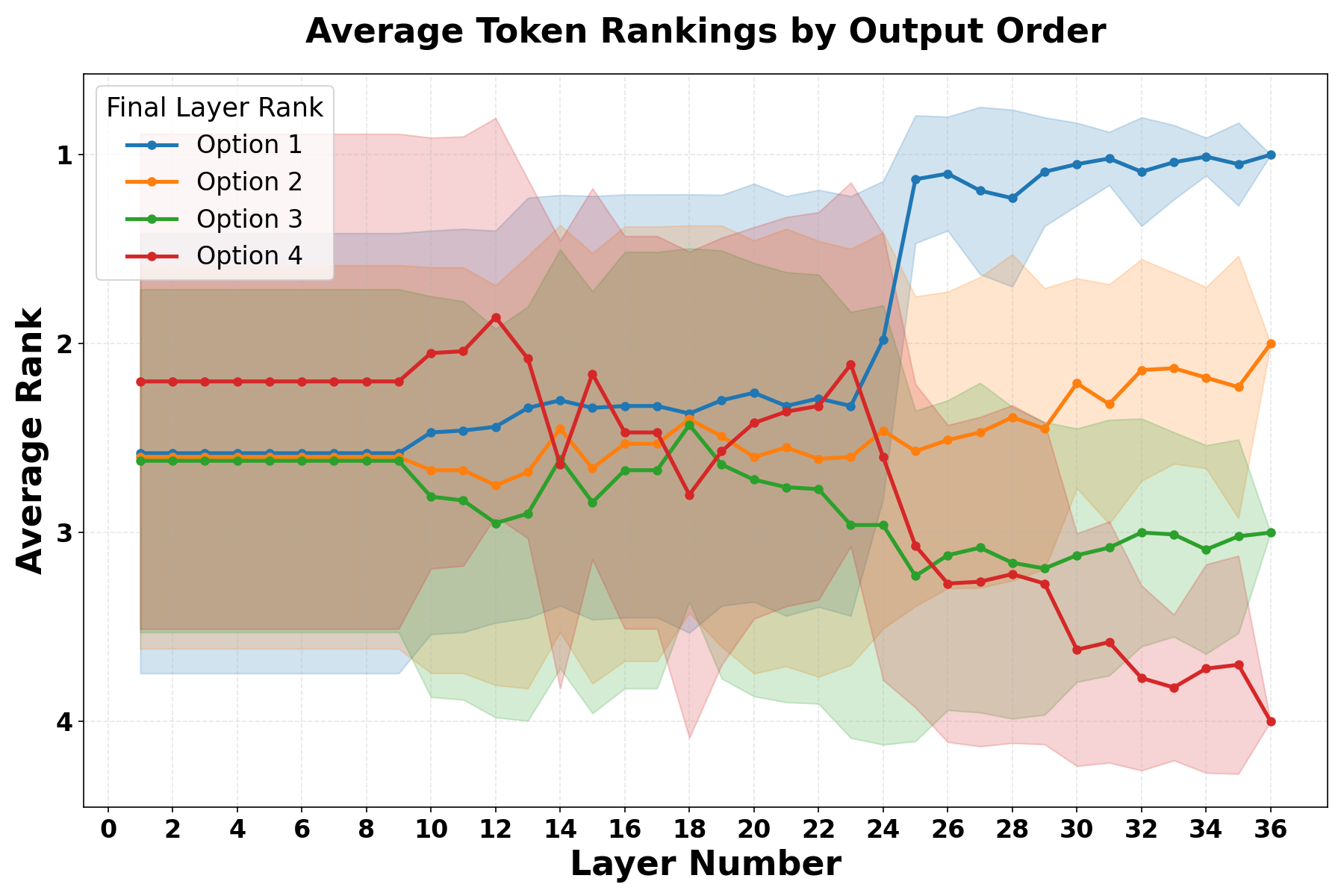}
\subcaption{Qwen (Arabic)}
\end{subfigure}
\begin{subfigure}{0.48\textwidth}
\centering
\includegraphics[width=\textwidth]{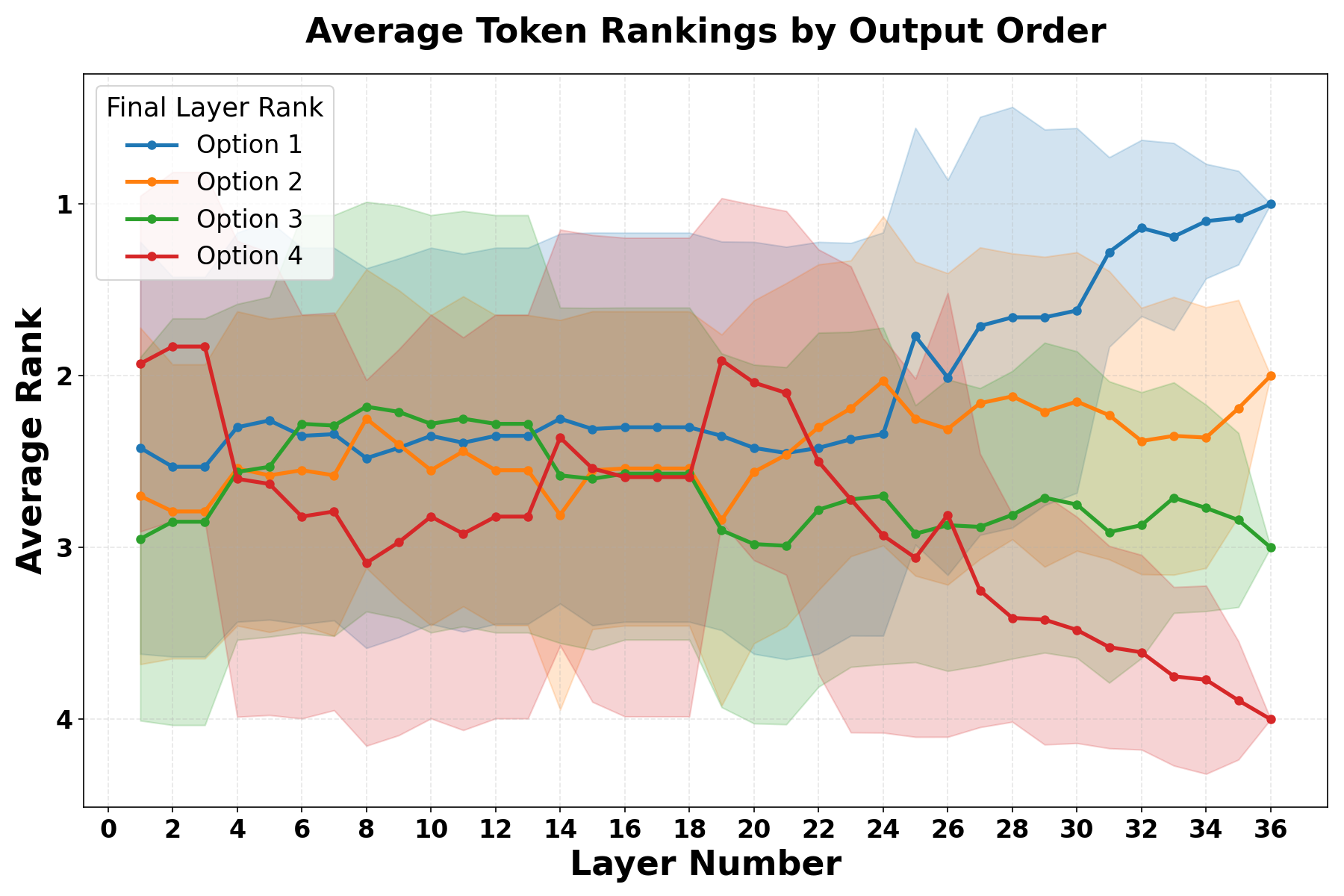}
\subcaption{Qwen (Roman)}
\end{subfigure}
\caption{Impact of option labeling schemes on token rankings across layers. Each subfigure shows aggregate token ranking plots for the Qwen model across alternate labeling schemes: Arabic numerals (1/2/3/4), and Roman numerals (i/ii/iii/iv), all evaluated on the QASC dataset. The plots reveal whether the location and prominence of the Hard Decision Layer (HDL) varies across different labeling schemes. Generally, the HDL manifests at a similar layer position regardless of labeling scheme, though its clarity differs: Alphabets (in Figure~\ref{fig:avg_rankings_qasc_qwen}) and Arabic numerals produce a well-pronounced HDL, while Roman numerals show a less distinct pattern.}
\label{fig:option_labels_aggregate}
\end{figure*}

\textbf{Finding 3: HDL prominence is affected by option labels used.} 
Previous experiments used alphabets (A/B/C/D) as option labels in the multiple-choice questions. To determine whether the choice of option labels affects the location or prominence of the HDL, we tested alternatives: arabic numerals (1/2/3/4) and roman numerals (i/ii/iii/iv). As shown in both Figure~\ref{fig:option_labels_aggregate} (also in Table~\ref{tab:hdl_label_variants} and Figure~\ref{fig:option_labels_aggregate_all} in Appendix~\ref{sec:supp-finding-2}), the HDL appears at the same position across all labeling schemes. However, the pattern differs in clarity: the HDL is well-pronounced for both arabic numerals and alphabets, whereas for roman numerals it is less distinct. The top-1 option remains stable post-HDL across all schemes, but other option probabilities show greater volatility with roman numerals compared to the other formats. The layer-wise accuracy plots in Figure~\ref{fig:option_labels_accuracy} - also in Appendix~\ref{sec:supp-finding-3} - reveal more pronounced differences. Arabic numerals exhibit the steepest accuracy jump at the HDL, with alphabets showing nearly identical behavior. In contrast, roman numerals display a more gradual accuracy increase beginning at the HDL. This might be indicative of MCQs in training data having more of alphabets and arabic numerals for option labels.

\subsection{Open-ended generation} 

We also briefly explore open-ended question answering by evaluating models on GSM8K math word problems. Residuals during generation of $n^{th}$ token were captured ($n = 10, 50, 100, 200$). Figure~\ref{fig:openended_generation} in Appendix~\ref{sec:plots-open-ended} displays average token rankings for the top 10 output tokens from the Qwen model; tokens not in the top-10 at intermediate layers appear blank. The HDL is less pronounced in these plots than in multiple-choice settings. However, knowing that layer 25 is the HDL for Qwen, we observe that the top-1 token stabilizes around this layer across all values of $n$, while other tokens show no clear pattern. This suggests the HDL may extend beyond multiple-choice answering to general token generation, though further investigation is needed.

\section{Discussion}
Our findings reveal interesting patterns in how language models process multiple-choice questions across their layers. At first glance, the observed behavior might be attributable to models storing dataset-specific information in particular layers, as suggested by prior work on model internals~\cite{meng2022locating}. However, this explanation appears unlikely for several reasons. The behavior remains consistent across diverse datasets, suggesting it is not tied to specific memorization of training data. Furthermore, the phenomenon occurs consistently across four different independently-trained models, making it improbable that every model independently chose to localize all dataset-specific knowledge for four different datasets in one layer. Additionally, the inclusion of SuperGPQA dataset, released after the Llama and Mistral models were trained, rules out dataset memorization as the primary driver, while still allowing for the remote possibility that relevant knowledge is localized within these layers.

A more compelling interpretation is that this behavior reflects the inherent structure of multiple-choice question answering itself. If this hypothesis holds, it would suggest that models may self-organize their computations in a task-specific manner, with later layers dedicated to selecting among the available options. Consistent with this view, our analysis of open-ended generation shows that while this decision-making pattern is less pronounced, it remains present. This aligns with recent findings in the literature showing that models identify likely candidate tokens in earlier layers and refine token distributions progressively through deeper layers~\cite{gupta2025llms}.

Our results have significant implications for model compression and efficiency. Since the HDL achieves accuracy comparable to the final layer, layers beyond the HDL can be pruned without sacrificing performance on multiple-choice tasks. Larger models like Llama demonstrate a shallower HDL position, enabling up to 43.75\% of layers to be removed while retaining competitive accuracy. This reveals an interesting interplay between model capacity and computational depth: a larger model with substantial layer pruning post-HDL may outperform a smaller, naturally efficient model with a deeper HDL. These results suggest that model size and HDL position are jointly important considerations for inference efficiency, providing a novel approach to trading off computational cost against model capacity. However, the findings also suggest a critical caveat to layer pruning: when prompt formatting differs, the final layers become essential for maintaining accuracy, and aggressive pruning post-HDL could substantially degrade performance.


\section{Limitations}
This study has several important limitations that should be considered when interpreting the results. First, our evaluation restricts models to direct answer selection without access to chain-of-thought reasoning, which is a more natural and often more effective way for models to approach complex problems. Additionally, our analysis focuses only on the canonical answer labels (A/B/C/D), but intermediate layers may activate alternative surface forms that correspond to the same option (e.g., ``b'', ``second'', ``B)'' and other variants), which are not accounted for in the current analysis. Furthermore, while our analysis includes some exploration of open-ended generation, this setting requires substantially more investigation to understand whether the patterns we observe in multiple-choice tasks generalize to free-form generation.  Future work should extend this analysis to chain-of-thought reasoning, more diverse generation tasks, account for answer form variations, and investigate the underlying mechanisms driving the observed layer-wise decision patterns.



\bibliography{references}

@article{hu2022lora,
  title={Lora: Low-rank adaptation of large language models.},
  author={Hu, Edward J and Shen, Yelong and Wallis, Phillip and Allen-Zhu, Zeyuan and Li, Yuanzhi and Wang, Shean and Wang, Liang and Chen, Weizhu and others},
  journal={Iclr},
  volume={1},
  number={2},
  pages={3},
  year={2022}
}

@article{gupta2025llms,
  title={How Do LLMs Use Their Depth?},
  author={Gupta, Akshat and Yeung, Jay and Anumanchipalli, Gopala and Ivanova, Anna},
  journal={arXiv preprint arXiv:2510.18871},
  year={2025}
}

@inproceedings{brown2020language,
    author    = {Tom B. Brown and Benjamin Mann and Nick Ryder and Melanie Subbiah and
                 Jared Kaplan and Prafulla Dhariwal and Arvind Neelakantan and
                 Pranav Shyam and Girish Sastry and Amanda Askell and Sandhini Agarwal and
                 Ariel Herbert-Voss and Gretchen Krueger and Tom Henighan and
                 Rewon Child and Aditya Ramesh and Daniel M. Ziegler and Jeffrey Wu and
                 Clemens Winter and Christopher Hesse and Mark Chen and Eric Sigler and
                 Mateusz Litwin and Scott Gray and Benjamin Chess and Jack Clark and
                 Christopher Berner and Sam McCandlish and Alec Radford and
                 Ilya Sutskever and Dario Amodei},
    title     = {Language Models are Few-Shot Learners},
    booktitle = {Advances in Neural Information Processing Systems 33 (NeurIPS)},
    year      = {2020},
    url       = {https://proceedings.neurips.cc/paper/2020/hash/1457c0d6bfcb4967418bfb8ac142f64a-Abstract.html}
}

@inproceedings{zhao2021calibrate,
    author    = {Tony Z. Zhao and Eric Wallace and Shi Feng and Dan Klein and Sameer Singh},
    title     = {Calibrate Before Use: Improving Few-Shot Performance of Language Models},
    booktitle = {Proceedings of the 38th International Conference on Machine Learning ({ICML})},
    series    = {Proceedings of Machine Learning Research},
    volume    = {139},
    pages     = {12697--12706},
    publisher = {PMLR},
    year      = {2021},
    url       = {http://proceedings.mlr.press/v139/zhao21c.html}
}

@article{elhage2021mathematical,
  title = {A Mathematical Framework for Transformer Circuits},
  author = {Elhage, Nelson and Nanda, Neel and Olsson, Catherine and Henighan, Tom and Joseph, Nicholas and Mann, Ben and Askell, Amanda and Bai, Yuntao and Chen, Anna and Conerly, Tom and others},
  journal = {Transformer Circuits Thread},
  year = {2021},
  url = {https://transformer-circuits.pub/2021/framework/index.html}
}

@inproceedings{hewitt2019structural,
  title = {A structural probe for finding syntax in word representations},
  author = {Hewitt, John and Manning, Christopher D.},
  booktitle = {Proceedings of the 2019 Conference of the North American Chapter of the Association for Computational Linguistics: Human Language Technologies},
  pages = {4129--4138},
  year = {2019}
}

@article{nostalgebraist2020logitlens,
  title={interpreting GPT: the logit lens},
  author={nostalgebraist},
  journal={LessWrong},
  year={2020},
  url={https://www.lesswrong.com/posts/AcKRB8wDpdaN6v6ru/interpreting-gpt-the-logit-lens}
}

@inproceedings{haviv2023understanding,
  title={Understanding transformer memorization recall through idioms},
  author={Haviv, Adi and Cohen, Ido and Gidron, Jacob and Schuster, Roei and Goldberg, Yoav and Geva, Mor},
  booktitle={Proceedings of the 17th Conference of the European Chapter of the Association for Computational Linguistics},
  pages={248--264},
  year={2023}
}

@inproceedings{dar2023analyzing,
  title = {Analyzing transformers in embedding space},
  author = {Dar, Guy and Geva, Mor and Gupta, Ankit and Berant, Jonathan},
  booktitle = {Proceedings of the 61st Annual Meeting of the Association for Computational Linguistics},
  pages = {16124--16170},
  year = {2023}
}

@inproceedings{geva2020transformer,
  title={Transformer feed-forward layers are key-value memories},
  author={Geva, Mor and Schuster, Roei and Berant, Jonathan and Levy, Omer},
  booktitle={Proceedings of the 2021 Conference on Empirical Methods in Natural Language Processing},
  pages={5484--5495},
  year={2021}
}

@inproceedings{meng2022locating,
  title = {Locating and editing factual associations in GPT},
  author = {Meng, Kevin and Bau, David and Andonian, Alex and Belinkov, Yonatan},
  booktitle = {Advances in Neural Information Processing Systems},
  volume = {35},
  pages = {17359--17372},
  year = {2022}
}

@inproceedings{din2023jump,
  title={Jump to conclusions: Short-cutting transformers with linear transformations},
  author={Din, Alexander Yom and Karidi, Taelin and Choshen, Leshem and Geva, Mor},
  booktitle={Proceedings of the 2024 Joint International Conference on Computational Linguistics, Language Resources and Evaluation (LREC-COLING 2024)},
  pages={9615--9625},
  year={2024}
}

@article{belrose2023eliciting,
  title={Eliciting latent predictions from transformers with the tuned lens},
  author={Belrose, Nora and Ostrovsky, Igor and McKinney, Lev and Furman, Zach and Smith, Logan and Halawi, Danny and Biderman, Stella and Steinhardt, Jacob},
  journal={arXiv preprint arXiv:2303.08112},
  year={2023}
}

@article{fan2024not,
  title={Not all layers of llms are necessary during inference},
  author={Fan, Siqi and Jiang, Xin and Li, Xiang and Meng, Xuying and Han, Peng and Shang, Shuo and Sun, Aixin and Wang, Yequan and Wang, Zhongyuan},
  journal={arXiv preprint arXiv:2403.02181},
  year={2024}
}

@inproceedings{geva2022transformer,
  title={Transformer feed-forward layers build predictions by promoting concepts in the vocabulary space},
  author={Geva, Mor and Caciularu, Avi and Wang, Kevin and Goldberg, Yoav},
  booktitle={Proceedings of the 2022 conference on empirical methods in natural language processing},
  pages={30--45},
  year={2022}
}

@article{lioubashevski2024looking,
  title={Looking beyond the top-1: Transformers determine top tokens in order},
  author={Lioubashevski, Daria and Schlank, Tomer and Stanovsky, Gabriel and Goldstein, Ariel},
  journal={arXiv preprint arXiv:2410.20210},
  year={2024}
}

@misc{bereska2024mechanistic,
  title = {Mechanistic interpretability for AI safety: A review},
  author = {Bereska, L. and Gavves, E.},
  year = {2024},
  howpublished = {arXiv preprint arXiv:2404.14082}
}

@article{hendrycks2021measuring,
  title = {Measuring massive multitask language understanding},
  author = {Hendrycks, Dan and Burns, Collin and Basart, Steven and Zou, Andy and Mazeika, Mantas and Song, Dawn and Steinhardt, Jacob},
  journal = {arXiv preprint arXiv:2009.03300},
  year = {2021}
}

@article{Jiang2023Mistral7,
  title={Mistral 7B},
  author={Albert Qiaochu Jiang and Alexandre Sablayrolles and Arthur Mensch and Chris Bamford and Devendra Singh Chaplot and Diego de Las Casas and Florian Bressand and Gianna Lengyel and Guillaume Lample and Lucile Saulnier and L{\'e}lio Renard Lavaud and Marie-Anne Lachaux and Pierre Stock and Teven Le Scao and Thibaut Lavril and Thomas Wang and Timoth{\'e}e Lacroix and William El Sayed},
  journal={ArXiv},
  year={2023},
  volume={abs/2310.06825},
  url={https://api.semanticscholar.org/CorpusID:263830494}
}

@article{grattafiori2024llama,
  title={The llama 3 herd of models},
  author={Grattafiori, Aaron and Dubey, Abhimanyu and Jauhri, Abhinav and Pandey, Abhinav and Kadian, Abhishek and Al-Dahle, Ahmad and Letman, Aiesha and Mathur, Akhil and Schelten, Alan and Vaughan, Alex and others},
  journal={arXiv preprint arXiv:2407.21783},
  year={2024}
}

@article{cobbe2021gsm8k,
  title={Training Verifiers to Solve Math Word Problems},
  author={Cobbe, Karl and Kosaraju, Vineet and Bavarian, Mohammad and Chen, Mark and Jun, Heewoo and Kaiser, Lukasz and Plappert, Matthias and Tworek, Jerry and Hilton, Jacob and Nakano, Reiichiro and Hesse, Christopher and Schulman, John},
  journal={arXiv preprint arXiv:2110.14168},
  year={2021}
}

@article{yang2025qwen3,
  title={Qwen3 technical report},
  author={Yang, An and Li, Anfeng and Yang, Baosong and Zhang, Beichen and Hui, Binyuan and Zheng, Bo and Yu, Bowen and Gao, Chang and Huang, Chengen and Lv, Chenxu and others},
  journal={arXiv preprint arXiv:2505.09388},
  year={2025}
}

@inproceedings{talmor-etal-2019-commonsenseqa,
    title = "{C}ommonsense{QA}: A Question Answering Challenge Targeting Commonsense Knowledge",
    author = "Talmor, Alon  and
      Herzig, Jonathan  and
      Lourie, Nicholas  and
      Berant, Jonathan",
    booktitle = "Proceedings of the 2019 Conference of the North {A}merican Chapter of the Association for Computational Linguistics: Human Language Technologies, Volume 1 (Long and Short Papers)",
    month = jun,
    year = "2019",
    address = "Minneapolis, Minnesota",
    publisher = "Association for Computational Linguistics",
    url = "https://aclanthology.org/N19-1421",
    doi = "10.18653/v1/N19-1421",
    pages = "4149--4158",
    archivePrefix = "arXiv",
    eprint        = "1811.00937",
    primaryClass  = "cs",
}

@article{allenai:qasc,
      author    = {Tushar Khot and Peter Clark and Michal Guerquin and Peter Jansen and Ashish Sabharwal},
      title     = {QASC: A Dataset for Question Answering via Sentence Composition},
      journal   = {arXiv:1910.11473v2},
      year      = {2020},
}

@article{wang2024mmlu,
  title={Mmlu-pro: A more robust and challenging multi-task language understanding benchmark},
  author={Wang, Yubo and Ma, Xueguang and Zhang, Ge and Ni, Yuansheng and Chandra, Abhranil and Guo, Shiguang and Ren, Weiming and Arulraj, Aaran and He, Xuan and Jiang, Ziyan and others},
  journal={Advances in Neural Information Processing Systems},
  volume={37},
  pages={95266--95290},
  year={2024}
}

@article{du2025supergpqa,
  title={Supergpqa: Scaling llm evaluation across 285 graduate disciplines},
  author={Du, Xinrun and Yao, Yifan and Ma, Kaijing and Wang, Bingli and Zheng, Tianyu and Zhu, King and Liu, Minghao and Liang, Yiming and Jin, Xiaolong and Wei, Zhenlin and others},
  journal={arXiv preprint arXiv:2502.14739},
  year={2025}
}

\newpage
\appendix

\newpage
\section*{Appendix Contents}
\begin{itemize}
  \item[\textbf{A}] Prompt, Models and Datasets \dotfill \pageref{sec:prompt-models-datasets}
  \item[\textbf{B}] LoRA Hyperparameter Details \dotfill \pageref{sec:hyperparams}
  \item[\textbf{C}] Supplementary Results for Finding 1 \dotfill \pageref{sec:supp-finding-1}
  \item[\textbf{D}] Supplementary Results for Finding 2 \dotfill \pageref{sec:supp-finding-2}
  \item[\textbf{E}] Supplementary Results for Finding 3 \dotfill \pageref{sec:supp-finding-3}
  \item[\textbf{F}] Open-Ended Generation \dotfill \pageref{sec:plots-open-ended}
\end{itemize}

\newpage

\section{Prompt, Models and Datasets}\label{sec:prompt-models-datasets}
Figure~\ref{fig:prompt_example} shows the prompt template used for the multiple-choice question answering used in experiments. Tables~\ref{tab:models} and \ref{tab:datasets} provide more information about the models and datasets used respectively.

\begin{figure}[!h]
\centering
\begin{tcolorbox}[colback=white,colframe=black,width=\linewidth]
You are given a question and some options. Output the correct option letter only and nothing else.

\textless question\textgreater 

For the two linear equations 2 * x + 3 * y = 10 and 4 * x + 4 * y = 12 with variables x and y. Use cramer's rule to solve these two variables.

\textless /question\textgreater

\textless options\textgreater

(A) [4, 1]

(B) [-2, 6]

(C) [3, 2]

(D) [-1, 4]

\textless /options\textgreater

The correct option is: (
\end{tcolorbox}
\caption{Prompt template used in the MCQA tasks.}
\label{fig:prompt_example}
\end{figure}

\begin{table*}[!h]
\caption{List of models evaluated in this study. Release dates are approximate and sourced from Hugging Face and arXiv.}
\label{tab:models}
\centering
\resizebox{\textwidth}{!}{
\begin{tabular}{lclp{4cm}}
\toprule
\textbf{Model} & \textbf{Parameter Count} & \textbf{Release Date} & \textbf{License} \\
\midrule
\href{https://huggingface.co/mistralai/Mistral-7B-Instruct-v0.3}{mistralai/Mistral-7B-Instruct-v0.3} & 7B & May 22, 2024 & Apache 2.0\\
\href{https://huggingface.co/meta-llama/Llama-3.1-8B-Instruct}{meta-llama/Llama-3.1-8B-Instruct} & 8B & July 23, 2024 & Llama 3.1 Community License Agreement\\
\href{https://huggingface.co/ibm-granite/granite-3.3-2b-instruct}{ibm-granite/granite-3.3-2b-instruct} & 2B & April 16, 2025 & Apache 2.0\\
\href{https://huggingface.co/Qwen/Qwen3-4B-Instruct-2507}{Qwen/Qwen3-4B-Instruct-2507} & 4B & August 5, 2025 & Apache 2.0\\
\bottomrule
\end{tabular}
}
\end{table*}

\begin{table*}[!h]
\caption{Summary of datasets used in this study. Release dates are approximate and sourced from Hugging Face and arXiv.}
\label{tab:datasets}
\centering
\resizebox{\textwidth}{!}{
\begin{tabular}{lp{6cm}lp{4cm}}
\toprule
\textbf{Dataset} & \textbf{Description} & \textbf{Release Date} & \textbf{License} \\
\midrule
\textbf{\href{https://huggingface.co/datasets/tau/commonsense_qa}{CommonsenseQA}} & Benchmark for evaluating commonsense reasoning capabilities through multiple-choice questions. & November 2, 2018 & MIT\\
\textbf{\href{https://huggingface.co/datasets/allenai/qasc}{QASC}} & Question-answering dataset requiring compositional reasoning and multi-step inference across multiple facts. & October 21, 2019 & Creative Commons Attribution 4.0\\

\textbf{\href{https://huggingface.co/datasets/TIGER-Lab/MMLU-Pro}{MMLU-Pro}} & Enhanced MMLU benchmark covering STEM, social sciences, and humanities with challenging multiple-choice questions. & May 8, 2024 & MIT\\
\textbf{\href{https://huggingface.co/datasets/m-a-p/SuperGPQA}{SuperGPQA}} & Graduate-level question-answering benchmark covering advanced concepts in physics, chemistry, and biology. & February 20, 2025 & Open Data Commons License Attribution\\
\bottomrule
\end{tabular}
}
\end{table*}

\section{LoRA Hyperparameter Details}\label{sec:hyperparams}

\paragraph{Training Hyperparameters}
Hyperparameters for all LoRA model trainings are set as follows:
\begin{itemize}
  \item Learning rate: $2 \times 10^{-4}$
  \item Maximum prompt length: $1024$
  \item Batch size: $64$
  \item LoRA rank: $32$
\end{itemize}

\paragraph{Implementation}
We use the Tinker library for implementing LoRA fine-tuning.

\paragraph{Dataset-specific Sample Counts}
The number of samples used for fine-tuning varies by dataset:
\begin{itemize}
  \item CommonsenseQA: $9000$ samples
  \item MMLU-Pro: $10000$ samples
  \item QASC: $8000$ samples
  \item SuperGPQA: $10000$ samples
\end{itemize}

\newpage
\section{Supplementary Results for Finding 1}\label{sec:supp-finding-1}
\begin{figure*}[!h]
	\centering
	\begin{subfigure}[b]{0.48\textwidth}
		\centering
		\includegraphics[width=\textwidth]{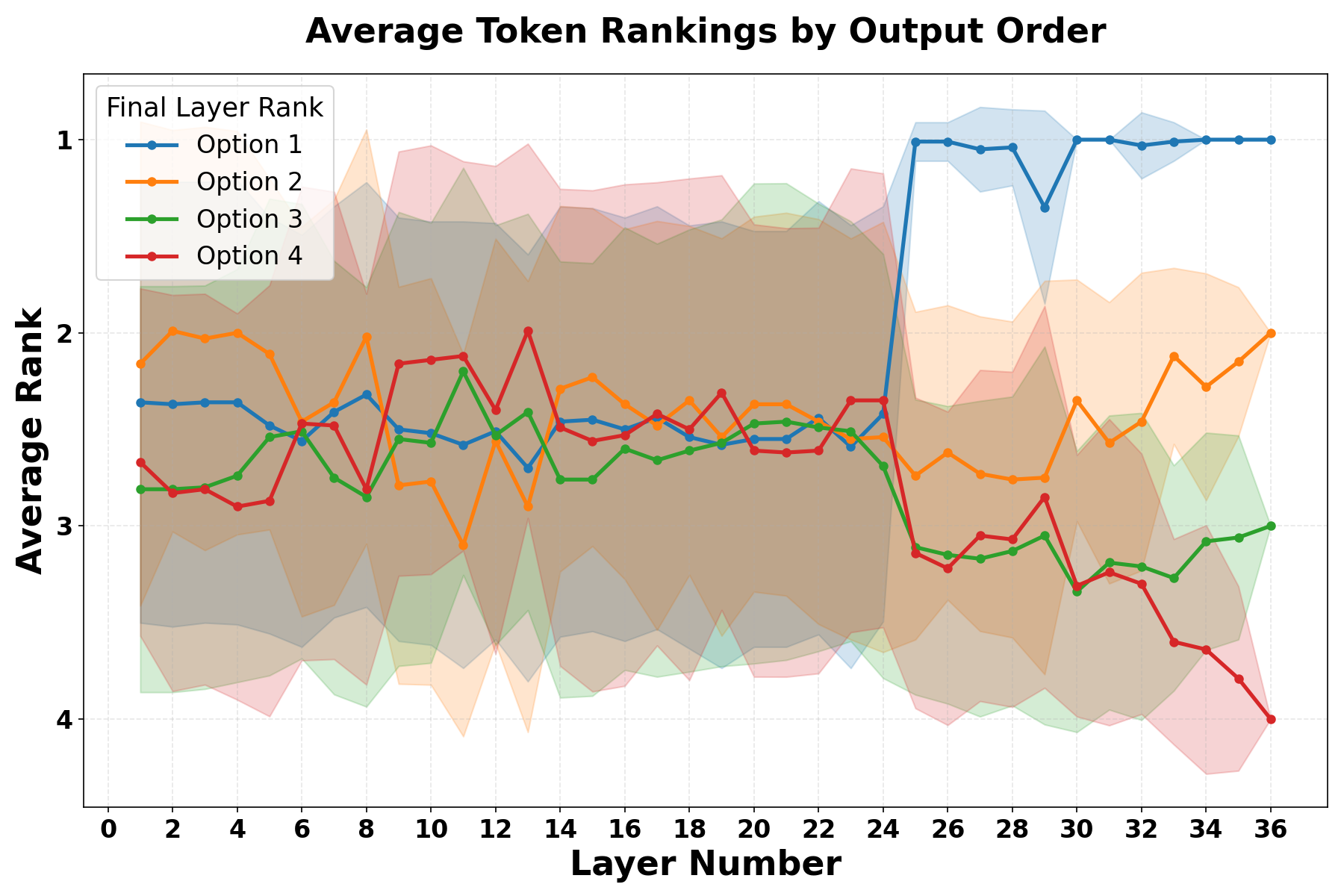}
		\caption{CommonsenseQA}
		\label{fig:hdl_commonsenseqa}
	\end{subfigure}
	\hfill
	\begin{subfigure}[b]{0.48\textwidth}
		\centering
		\includegraphics[width=\textwidth]{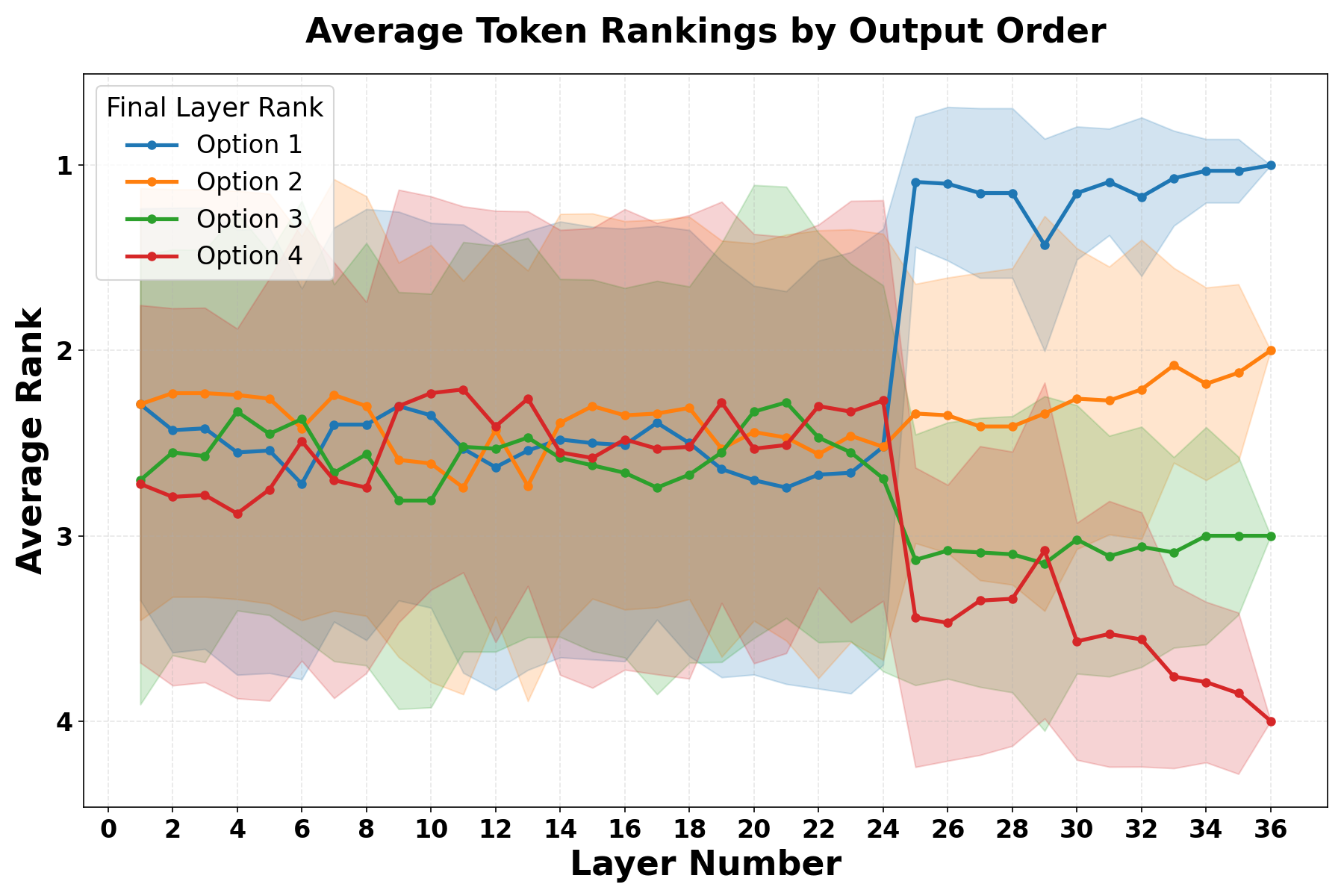}
		\caption{MMLU-Pro}
		\label{fig:hdl_mmlu_pro}
	\end{subfigure}
	\\[1em]
	\begin{subfigure}[b]{0.48\textwidth}
		\centering
		\includegraphics[width=\textwidth]{images/plots_for_paper/qasc/qwen/aggregate_plots/Correct/aggregate_overall_by_output_order.png}
		\caption{QASC}
		\label{fig:hdl_qasc}
	\end{subfigure}
	\hfill
	\begin{subfigure}[b]{0.48\textwidth}
		\centering
		\includegraphics[width=\textwidth]{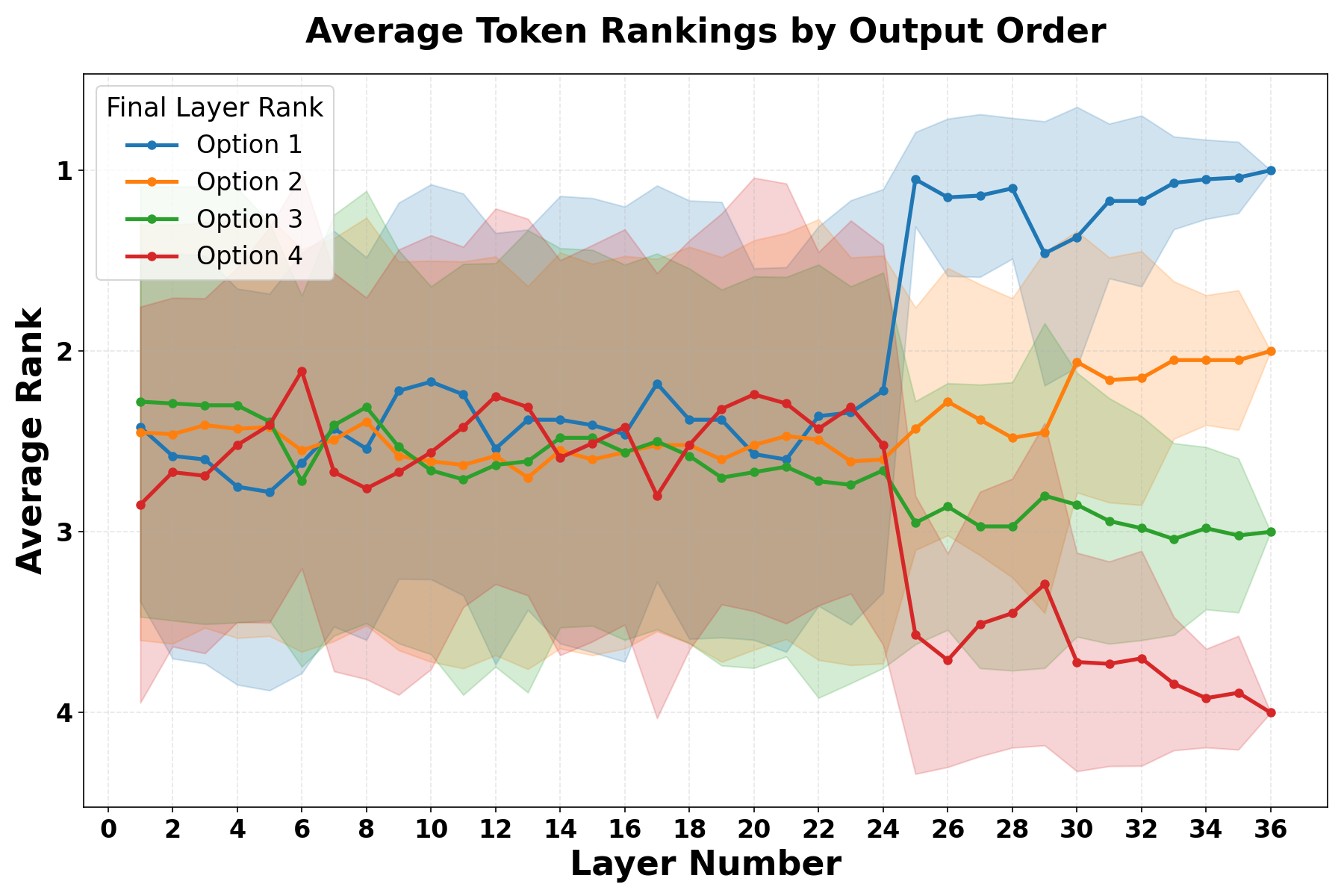}
		\caption{SuperGPQA}
		\label{fig:hdl_supergpqa}
	\end{subfigure}
	\caption{Average token rankings plots of the Qwen model when evaluated on different datasets.}
	\label{fig:hdl_vs_dataset_qwen}
\end{figure*}

\begin{figure*}[!h]
	\centering
	\begin{subfigure}[b]{0.48\textwidth}
		\centering
		\includegraphics[width=\textwidth]{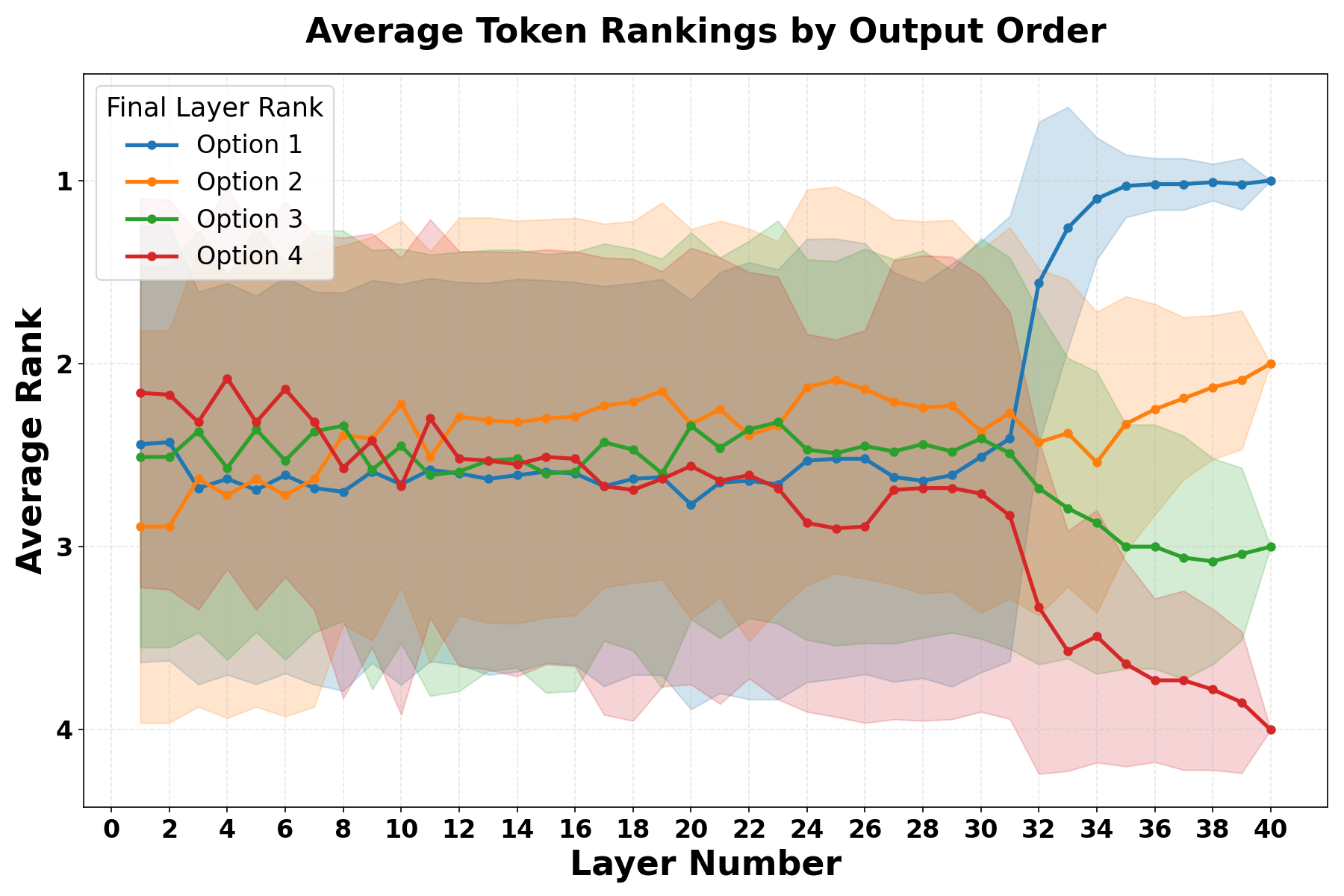}
		\caption{Granite}
		\label{fig:hdl_granite}
	\end{subfigure}
	\hfill
	\begin{subfigure}[b]{0.48\textwidth}
		\centering
		\includegraphics[width=\textwidth]{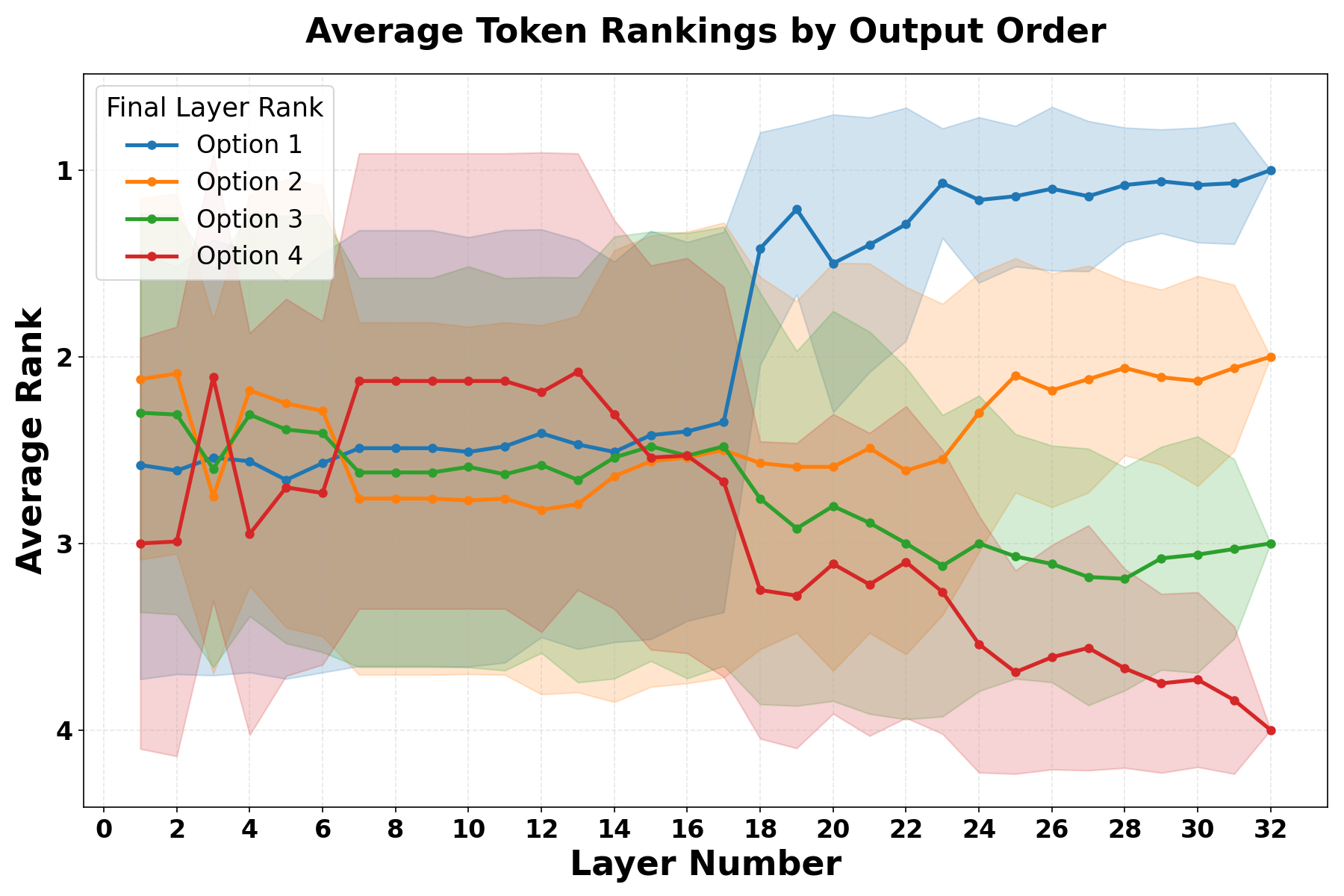}
		\caption{Llama}
		\label{fig:hdl_llama}
	\end{subfigure}
	\\[1em]
	\begin{subfigure}[b]{0.48\textwidth}
		\centering
		\includegraphics[width=\textwidth]{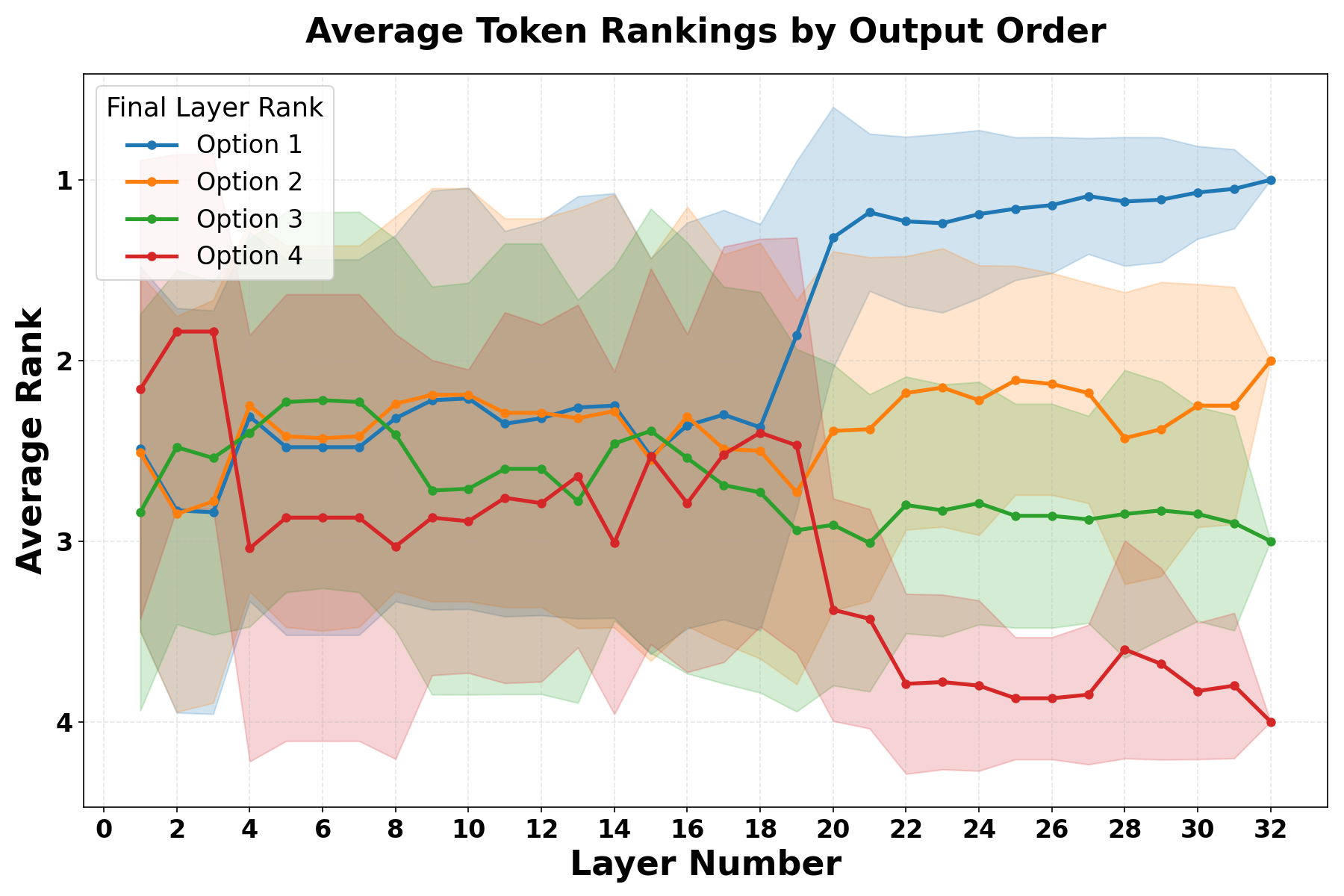}
		\caption{Mistral}
		\label{fig:hdl_mistral}
	\end{subfigure}
	\hfill
	\begin{subfigure}[b]{0.48\textwidth}
		\centering
		\includegraphics[width=\textwidth]{images/plots_for_paper/qasc/qwen/aggregate_plots/Correct/aggregate_overall_by_output_order.png}
		\caption{Qwen}
		\label{fig:hdl_qwen}
	\end{subfigure}
	\caption{Average token rankings plots across different models when evaluated on the QASC dataset.}
	\label{fig:hdl_vs_model_qasc}
\end{figure*}

Figure~\ref{fig:hdl_vs_dataset_qwen} shows the average token rankings through the layers for the Qwen model on all the four datasets evaluated. Figure~\ref{fig:hdl_vs_model_qasc} shows the average token rankings through the layers for all four models on the QASC dataset.

\section{Supplementary Results for Finding 2}\label{sec:supp-finding-2}
\begin{figure*}[!h]
	\centering
	\begin{subfigure}[b]{0.48\textwidth}
		\centering
		\includegraphics[width=\textwidth]{images/plots_for_paper/qasc/qwen/aggregate_plots/Correct/aggregate_overall_by_output_order.png}
		\caption{QASC Qwen (Base)}
		\label{fig:hdl_qasc_qwen_base}
	\end{subfigure}
	\hfill
	\begin{subfigure}[b]{0.48\textwidth}
		\centering
		\includegraphics[width=\textwidth]{images/plots_for_paper/mmlu_pro/qwen/aggregate_plots/Correct/aggregate_overall_by_output_order.png}
		\caption{MMLU-Pro Qwen (Base)}
		\label{fig:hdl_mmlu_qwen_base}
	\end{subfigure}
	\\[0.7em]
	\begin{subfigure}[b]{0.48\textwidth}
		\centering
		\includegraphics[width=\textwidth]{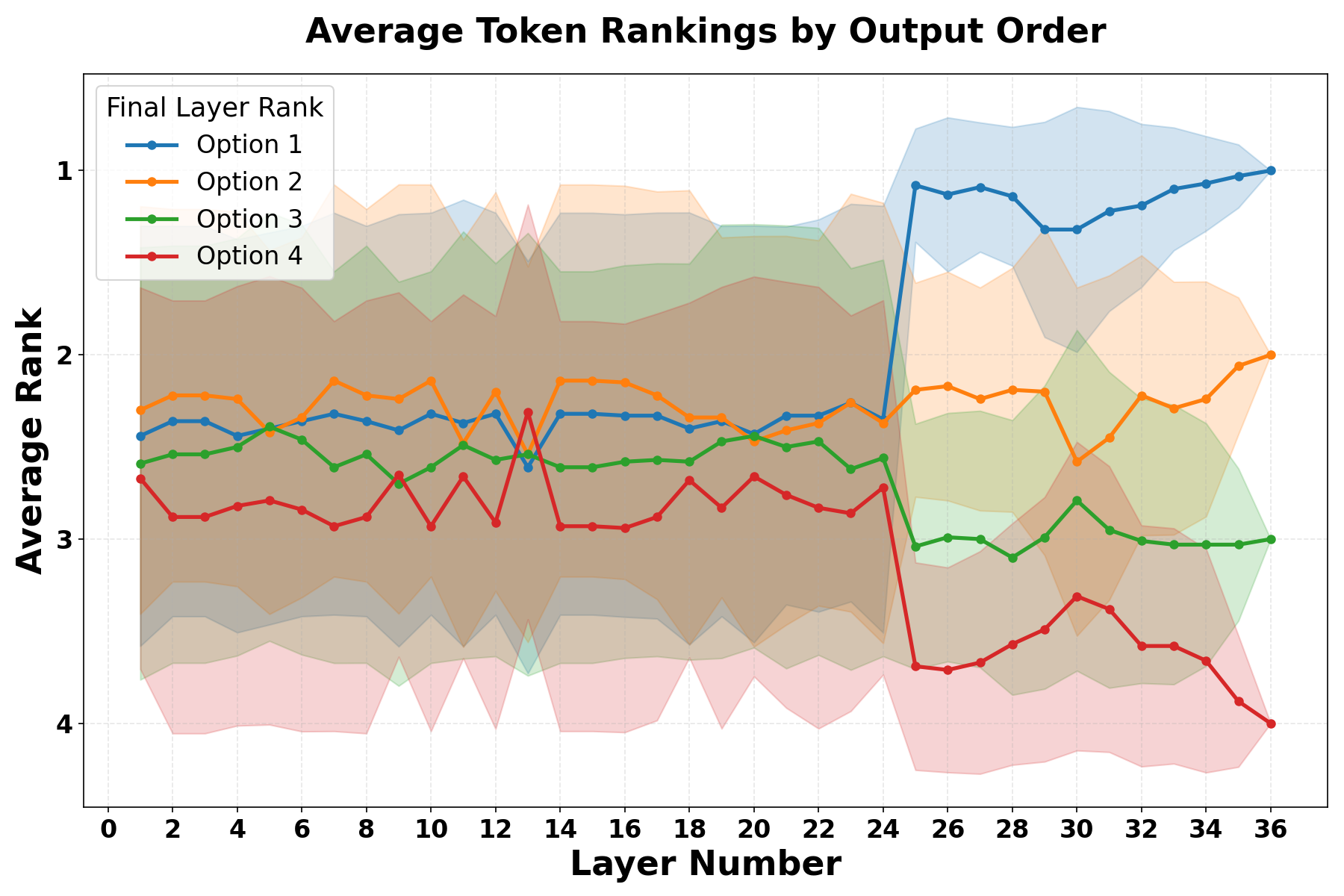}
		\caption{QASC Qwen (Ft)}
		\label{fig:hdl_qasc_qwen_ft}
	\end{subfigure}
	\hfill
	\begin{subfigure}[b]{0.48\textwidth}
		\centering
		\includegraphics[width=\textwidth]{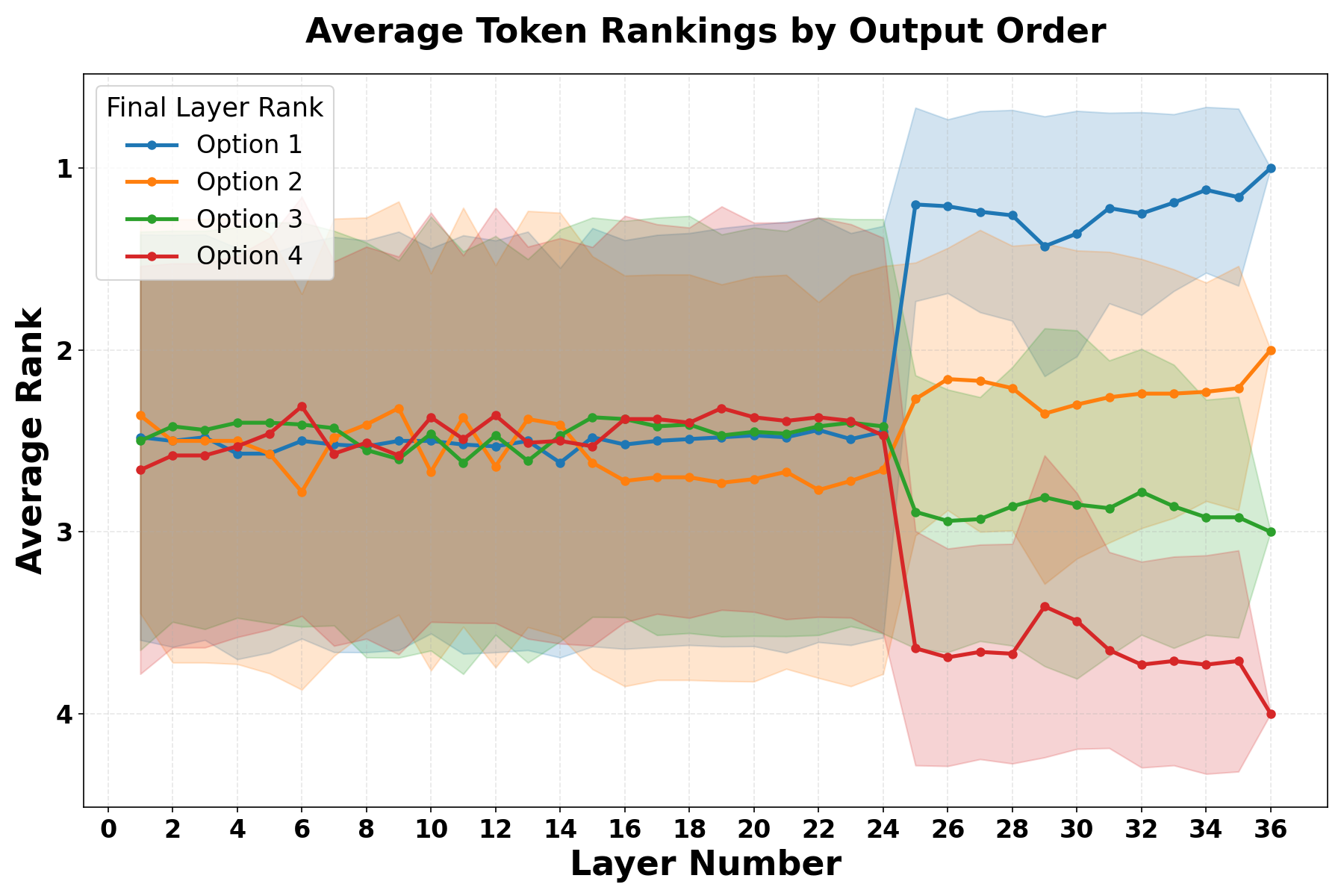}
		\caption{MMLU-Pro Qwen (Ft)}
		\label{fig:hdl_mmlu_qwen_ft}
	\end{subfigure}
	\caption{Average token rankings plots comparing base and finetuned (Ft) Qwen models across datasets.}
	\label{fig:hdl_qwen_finetuning}
\end{figure*}

\begin{figure*}[!h]
	\centering
	\begin{subfigure}[b]{0.48\textwidth}
		\centering
		\includegraphics[width=\textwidth]{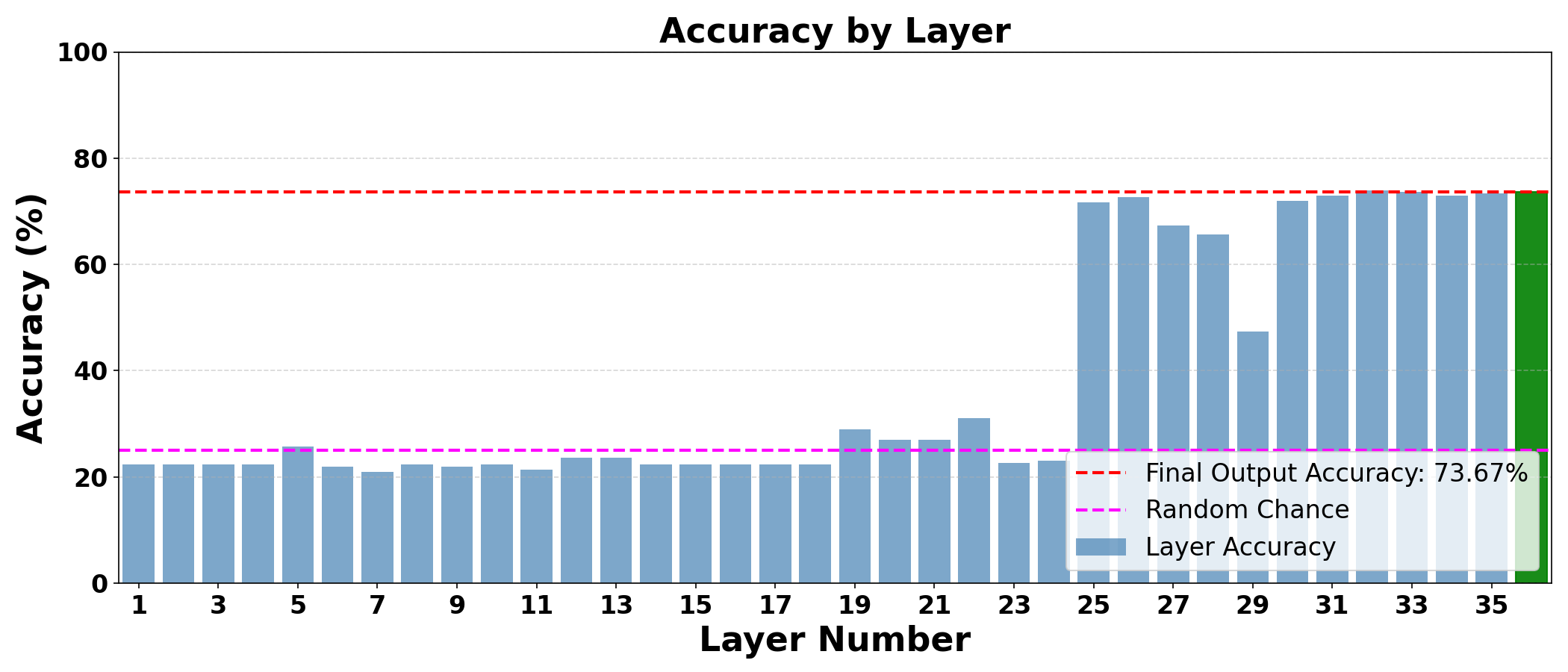}
		\caption{Qwen QASC \\(Base)}
		\label{fig:hdl_qwen_qasc_base}
	\end{subfigure}
	\hfill
	\begin{subfigure}[b]{0.48\textwidth}
		\centering
		\includegraphics[width=\textwidth]{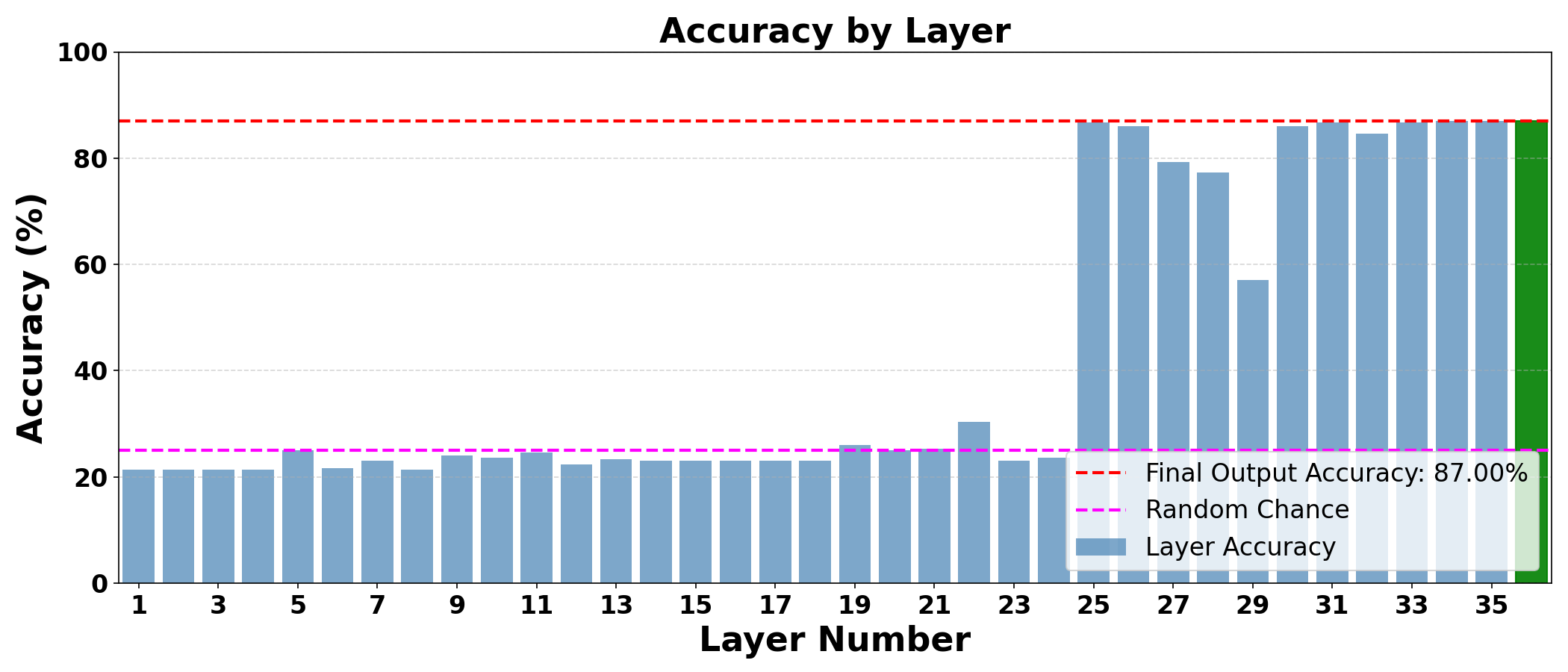}
		\caption{Qwen CommonsenseQA \\(Base)}
		\label{fig:hdl_qwen_csqa_base}
	\end{subfigure}
	\\[0.7em]
	\begin{subfigure}[b]{0.48\textwidth}
		\centering
		\includegraphics[width=\textwidth]{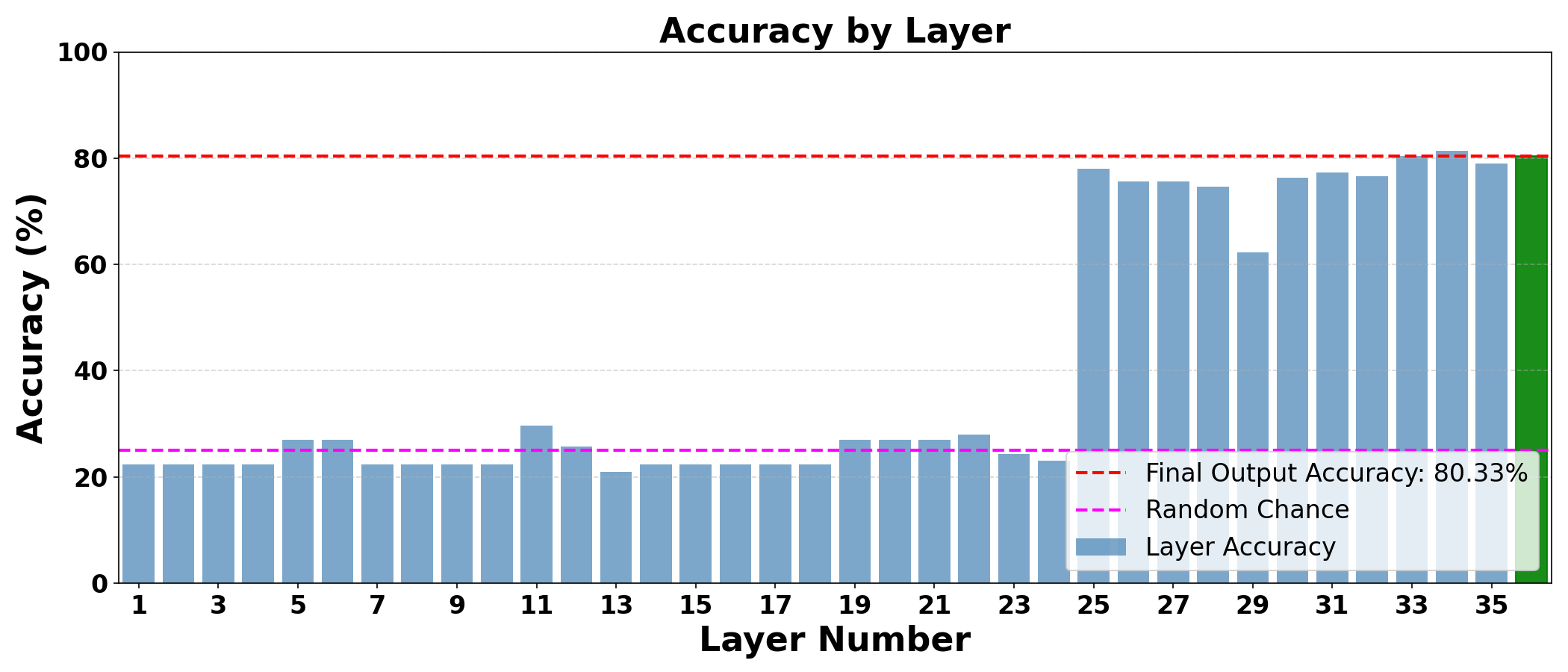}
		\caption{Qwen QASC \\(Ft)}
		\label{fig:hdl_qwen_qasc_ft}
	\end{subfigure}
	\hfill
	\begin{subfigure}[b]{0.48\textwidth}
		\centering
		\includegraphics[width=\textwidth]{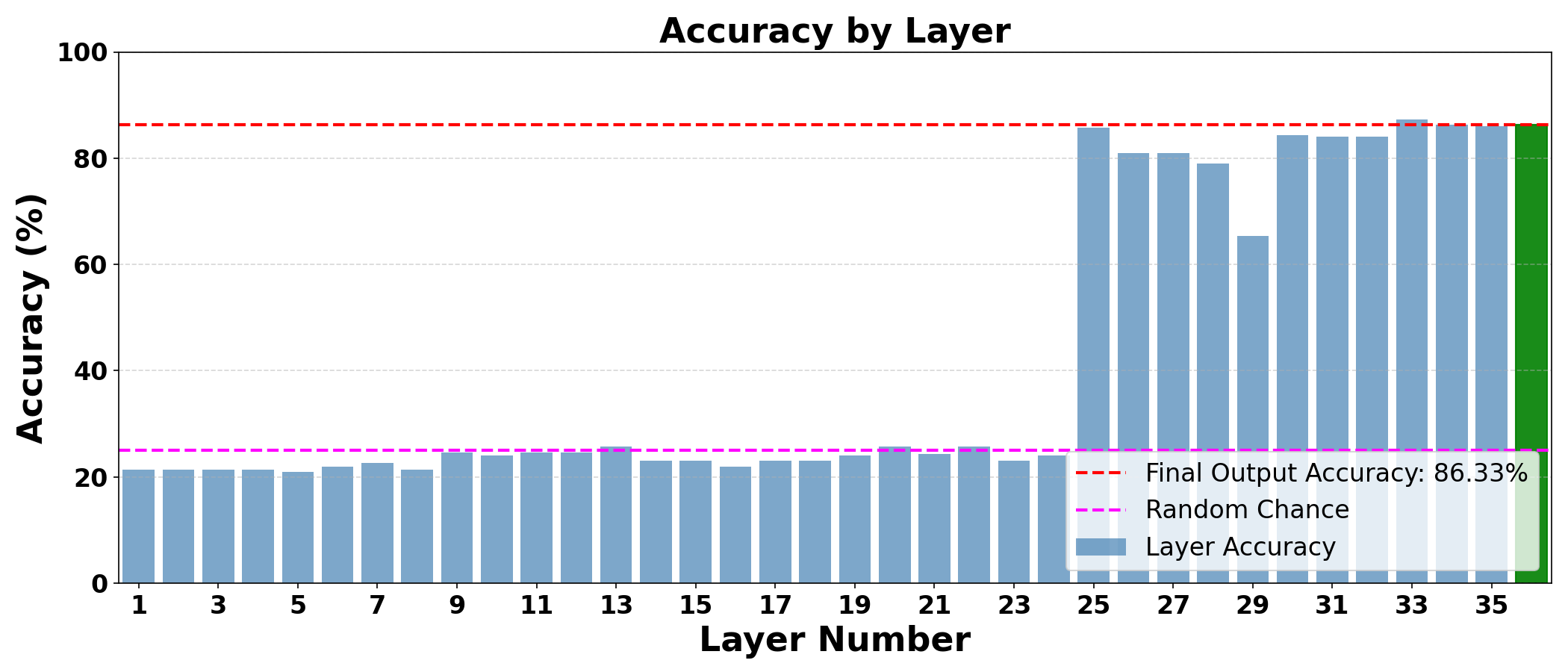}
		\caption{Qwen CommonsenseQA \\(Ft)}
		\label{fig:hdl_qwen_csqa_ft}
	\end{subfigure}
	\caption{Accuracy of intermediate layer outputs across model layers for base and finetuned (Ft) Qwen models.}
	\label{fig:hdl_qwen_accuracy}
\end{figure*}

\begin{figure*}[!h]
	\centering
	\begin{subfigure}[b]{0.48\textwidth}
		\centering
		\includegraphics[width=\textwidth]{images/plots_for_paper/qasc/llama/aggregate_plots/Correct/aggregate_overall_by_output_order.png}
		\caption{QASC Llama (Base)}
		\label{fig:hdl_qasc_llama_base}
	\end{subfigure}
	\hfill
	\begin{subfigure}[b]{0.48\textwidth}
		\centering
		\includegraphics[width=\textwidth]{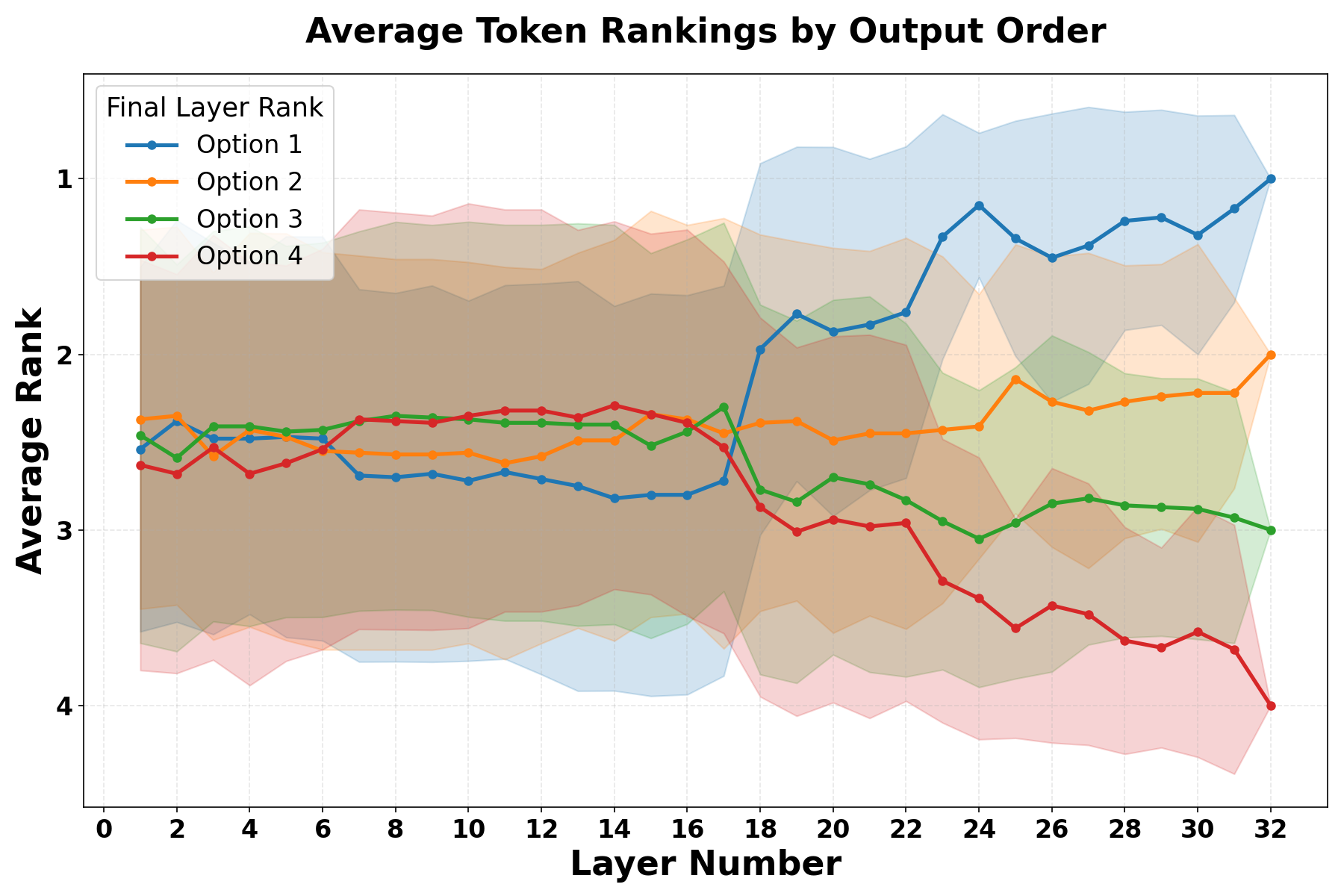}
		\caption{MMLU-Pro Llama (Base)}
		\label{fig:hdl_mmlu_llama_base}
	\end{subfigure}
	\\[0.7em]
	\begin{subfigure}[b]{0.48\textwidth}
		\centering
		\includegraphics[width=\textwidth]{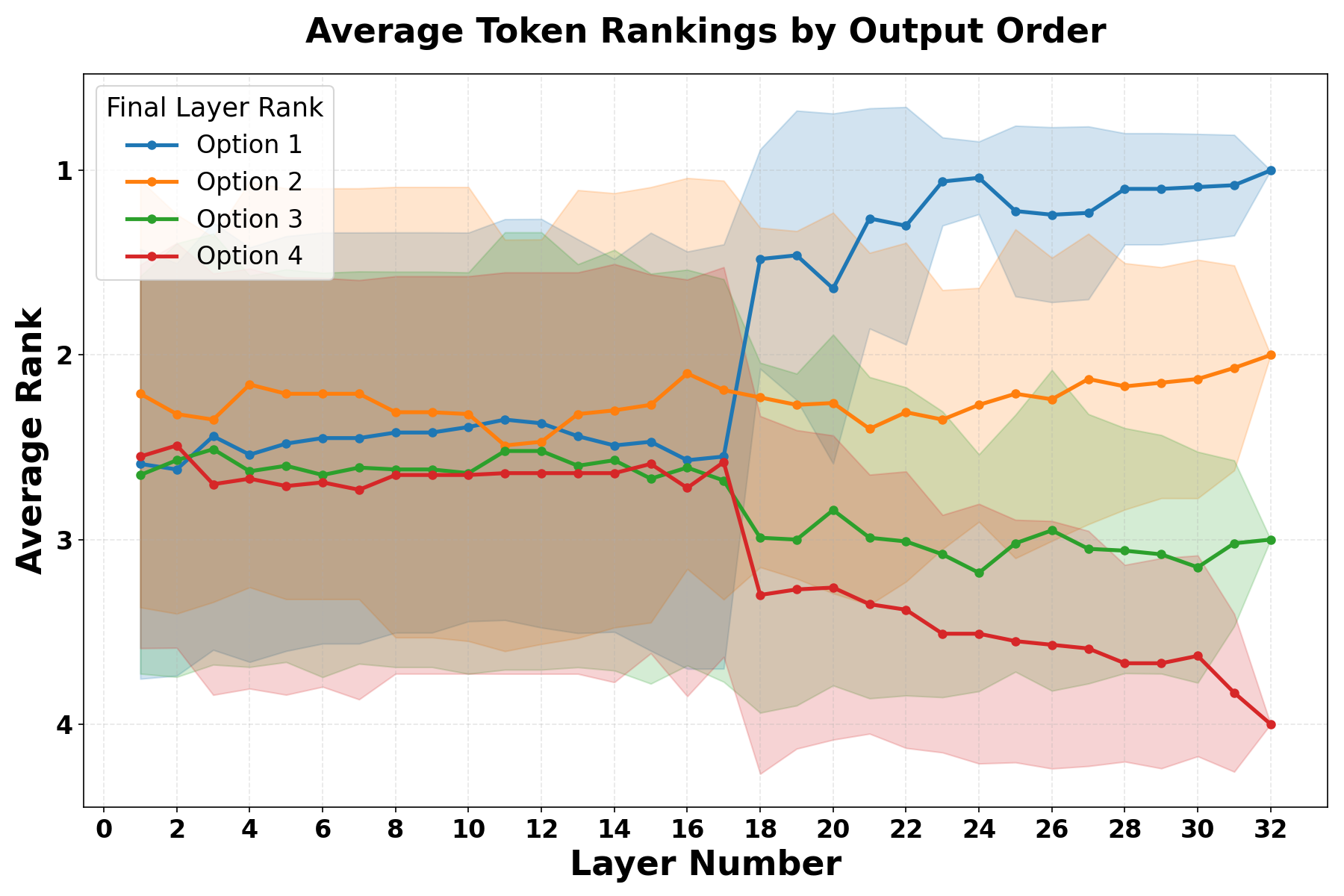}
		\caption{QASC Llama (Ft)}
		\label{fig:hdl_qasc_llama_ft}
	\end{subfigure}
	\hfill
	\begin{subfigure}[b]{0.48\textwidth}
		\centering
		\includegraphics[width=\textwidth]{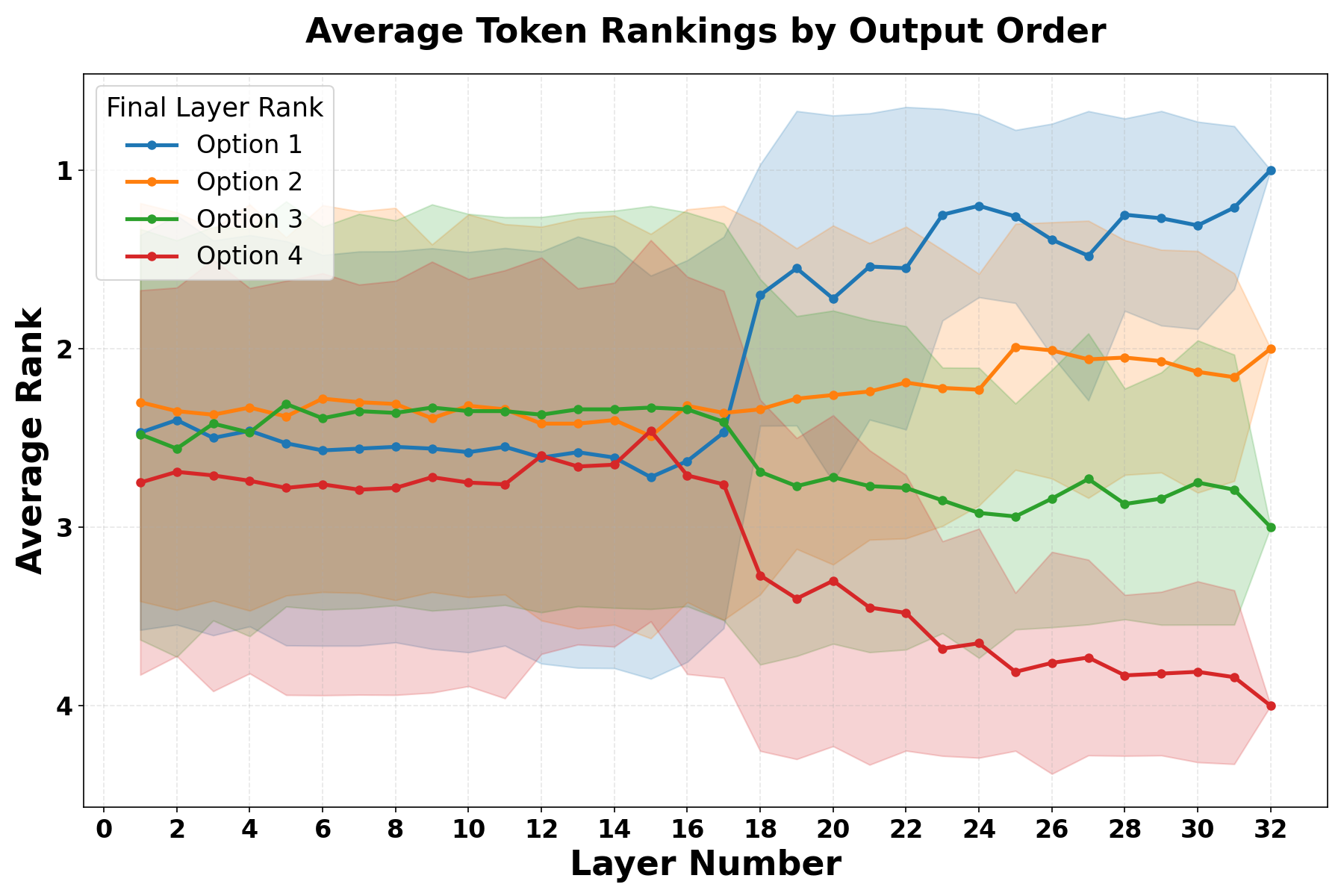}
		\caption{MMLU-Pro Llama (Ft)}
		\label{fig:hdl_mmlu_llama_ft}
	\end{subfigure}
	\caption{Average token rankings plots comparing base and finetuned (Ft) Llama models across datasets.}
	\label{fig:hdl_llama_finetuning}
\end{figure*}

\begin{figure*}[!h]
	\centering
	\begin{subfigure}[b]{0.48\textwidth}
		\centering
		\includegraphics[width=\textwidth]{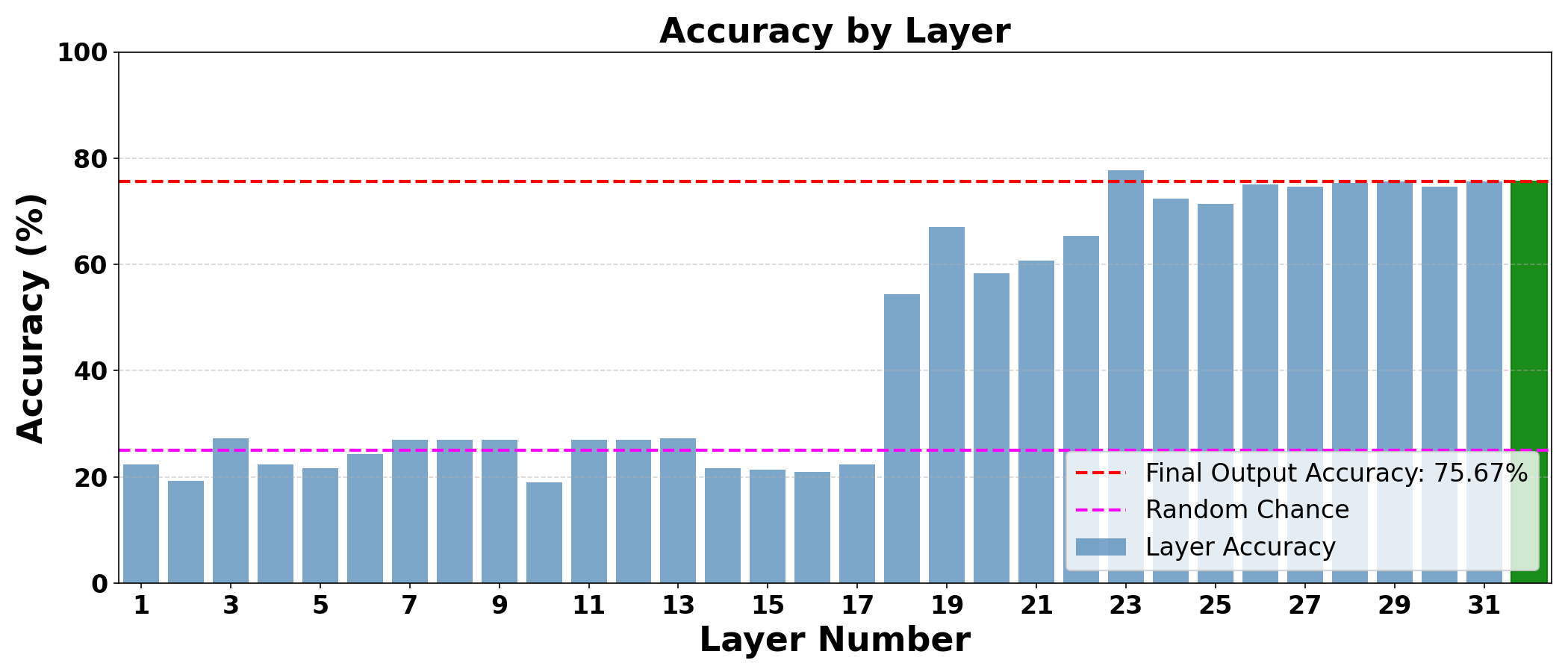}
		\caption{Llama QASC \\(Base)}
		\label{fig:hdl_llama_qasc_base}
	\end{subfigure}
	\hfill
	\begin{subfigure}[b]{0.48\textwidth}
		\centering
		\includegraphics[width=\textwidth]{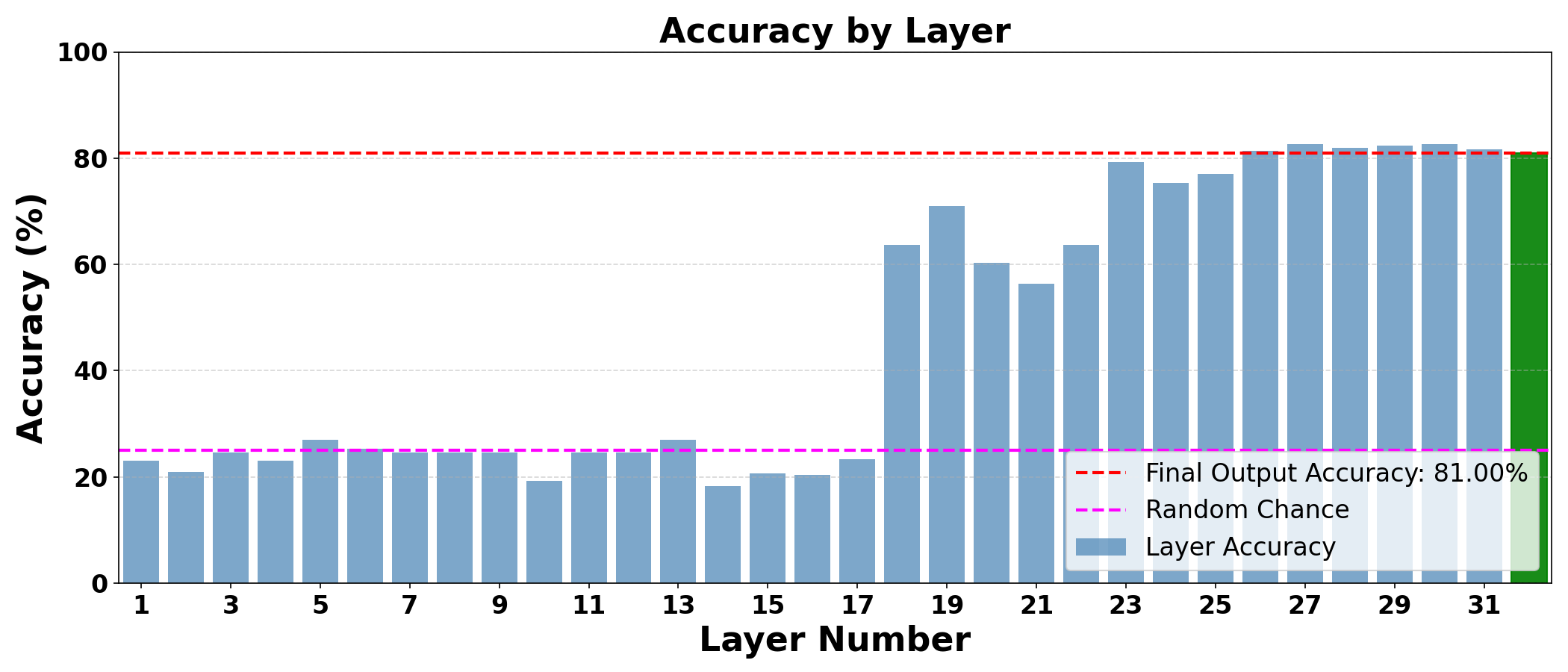}
		\caption{Llama CommonsenseQA \\(Base)}
		\label{fig:hdl_llama_csqa_base}
	\end{subfigure}
	\\[0.7em]
	\begin{subfigure}[b]{0.48\textwidth}
		\centering
		\includegraphics[width=\textwidth]{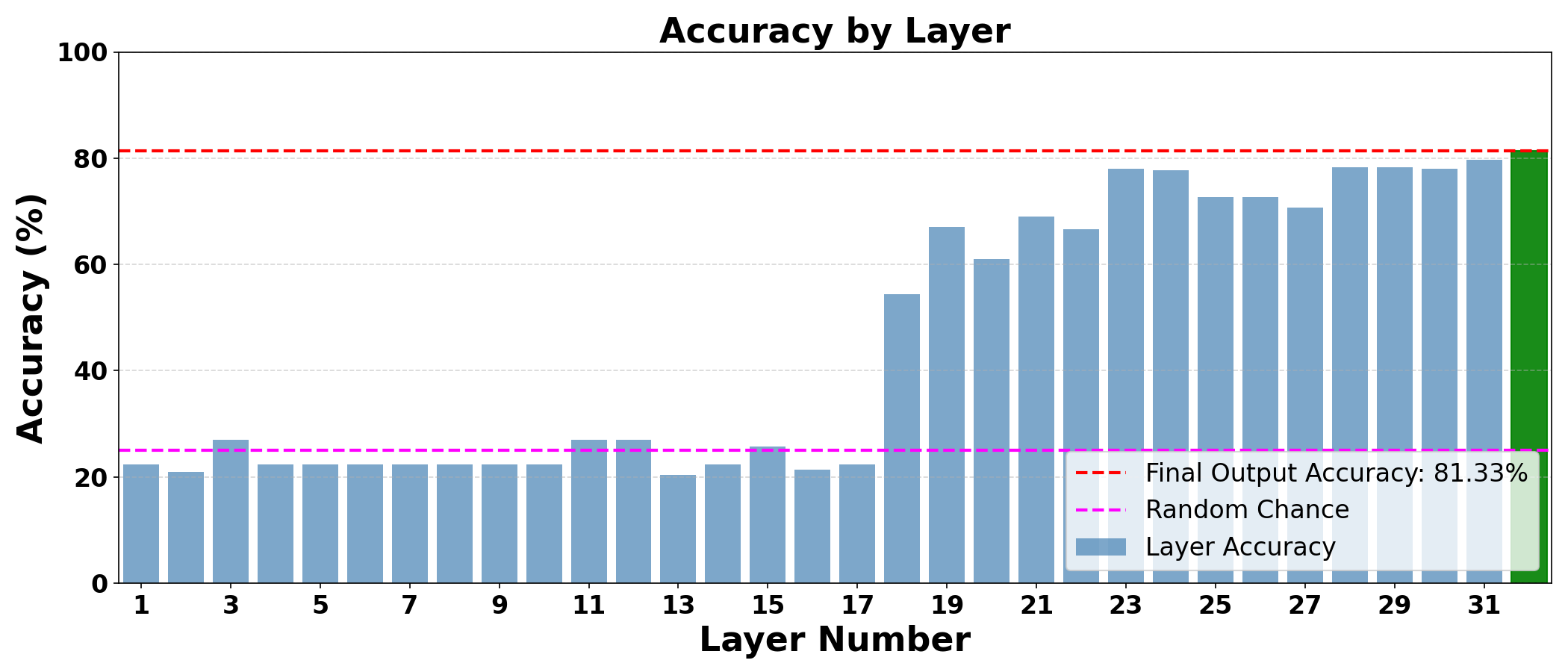}
		\caption{Llama QASC \\(Ft)}
		\label{fig:hdl_llama_qasc_ft}
	\end{subfigure}
	\hfill
	\begin{subfigure}[b]{0.48\textwidth}
		\centering
		\includegraphics[width=\textwidth]{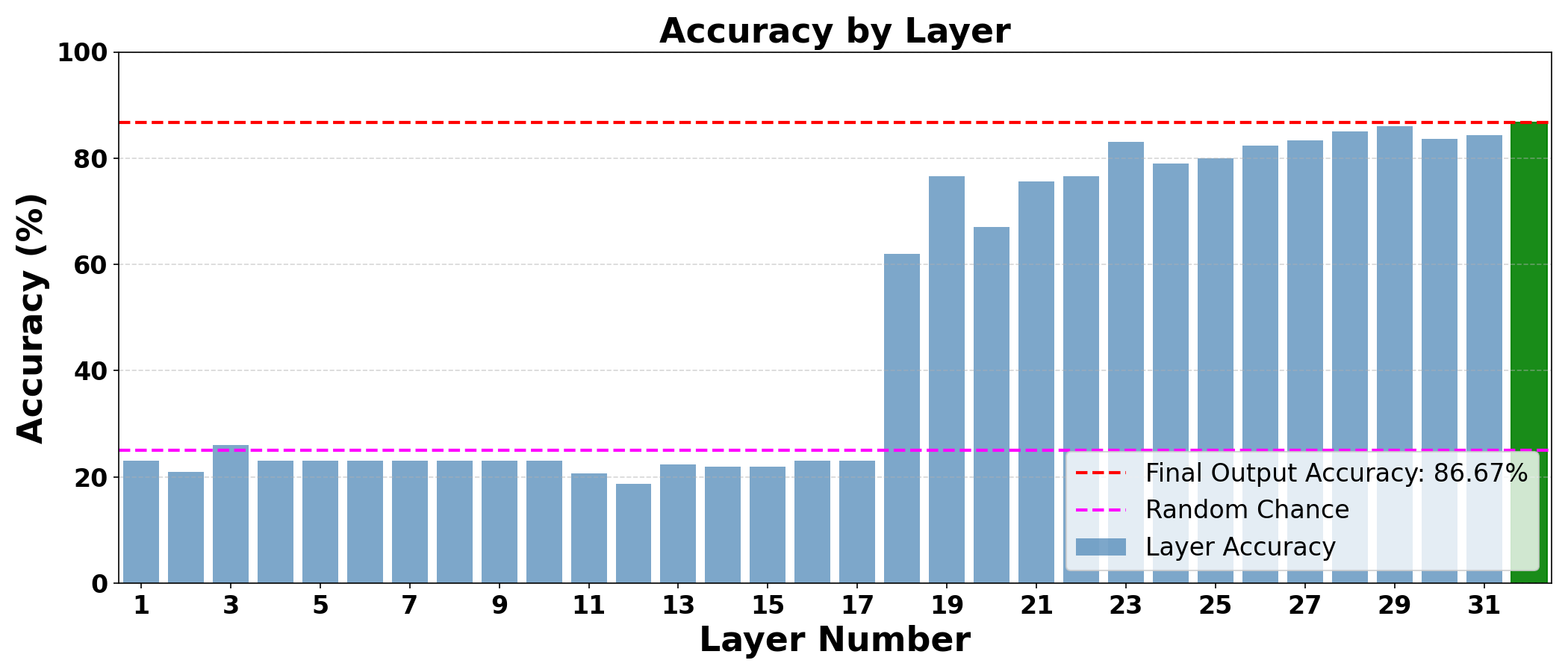}
		\caption{Llama CommonsenseQA \\(Ft)}
		\label{fig:hdl_llama_csqa_ft}
	\end{subfigure}
	\caption{Accuracy of intermediate layer outputs across model layers for base and finetuned (Ft) Llama models.}
	\label{fig:hdl_llama_accuracy}
\end{figure*}

Figure~\ref{fig:hdl_qwen_finetuning} and Figure~\ref{fig:hdl_qwen_accuracy} show the average token ranking and accuracy through the layers for the Qwen model in comparison to its finetuned versions on QASC and MMLU-Pro datasets, while Figure~\ref{fig:hdl_llama_finetuning} and Figure~\ref{fig:hdl_llama_accuracy} show the corresponding results for the Llama model.

\section{Supplementary Results for Finding 3}\label{sec:supp-finding-3}
\begin{figure*}[!h]
\centering
\begin{subfigure}{0.48\textwidth}
\centering
\includegraphics[width=\textwidth]{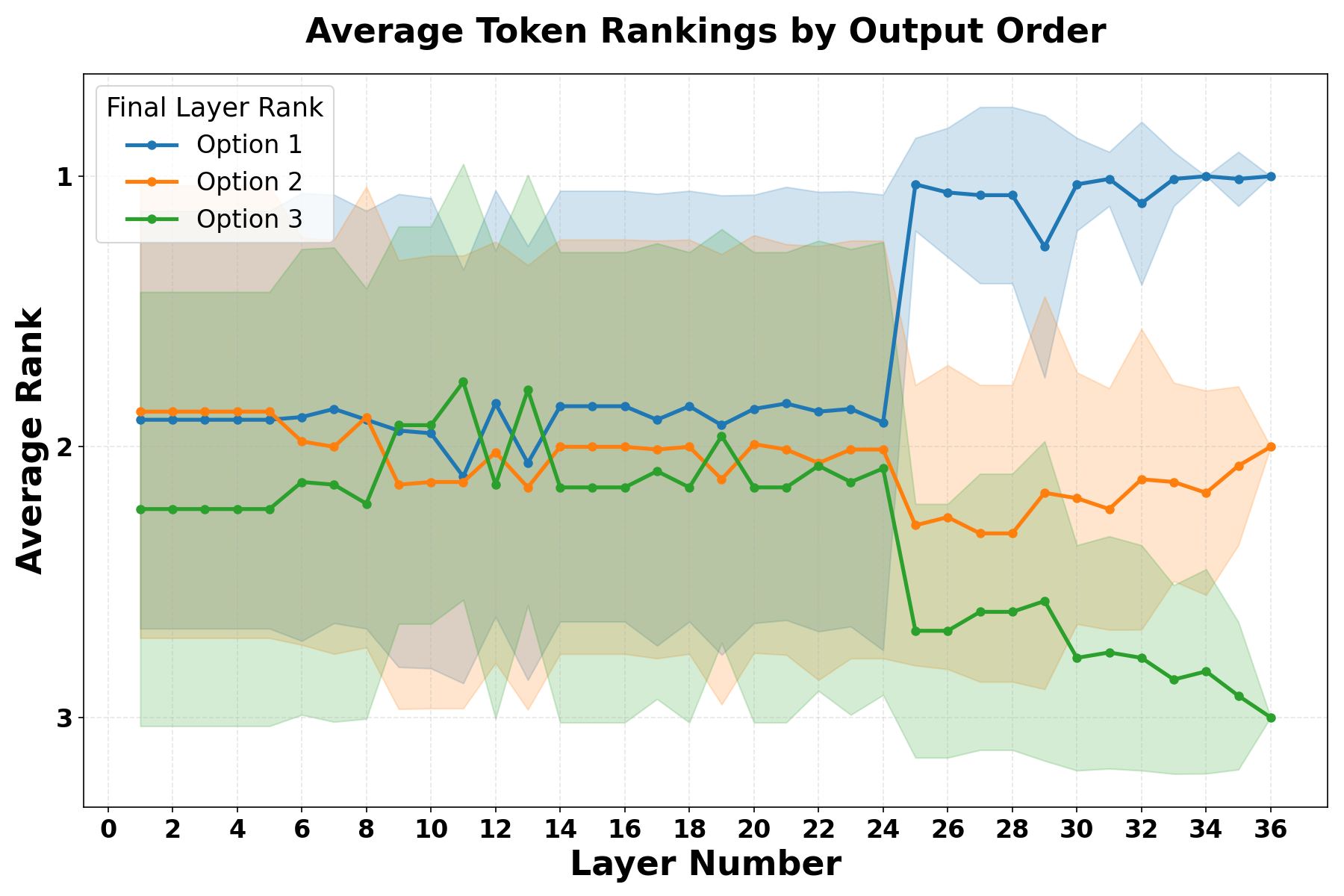}
\subcaption{Qwen (3 Options)}
\end{subfigure}
\hfill
\begin{subfigure}{0.48\textwidth}
\centering
\includegraphics[width=\textwidth]{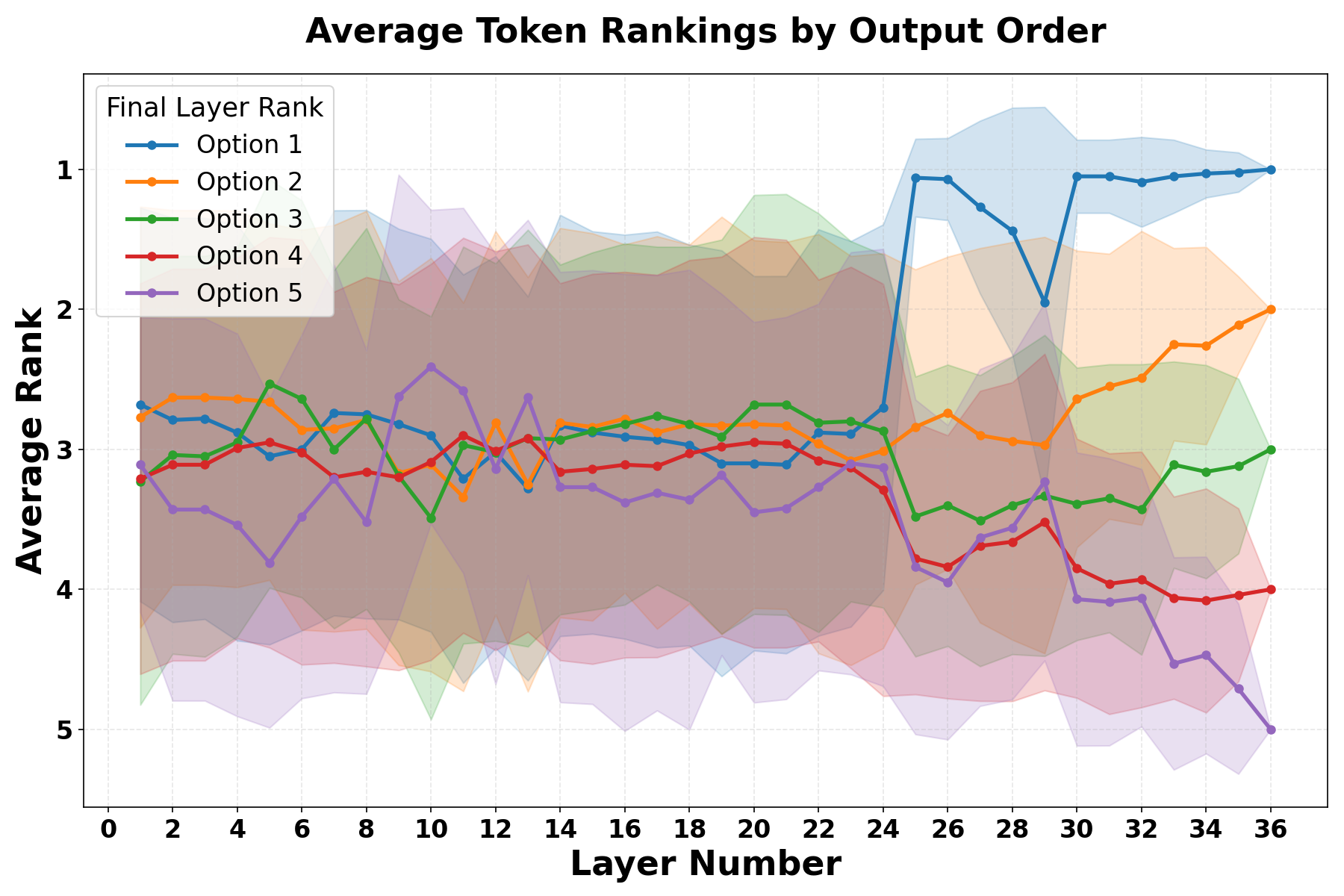}
\subcaption{Qwen (5 Options)}
\end{subfigure}
\\
\begin{subfigure}{0.48\textwidth}
\centering
\includegraphics[width=\textwidth]{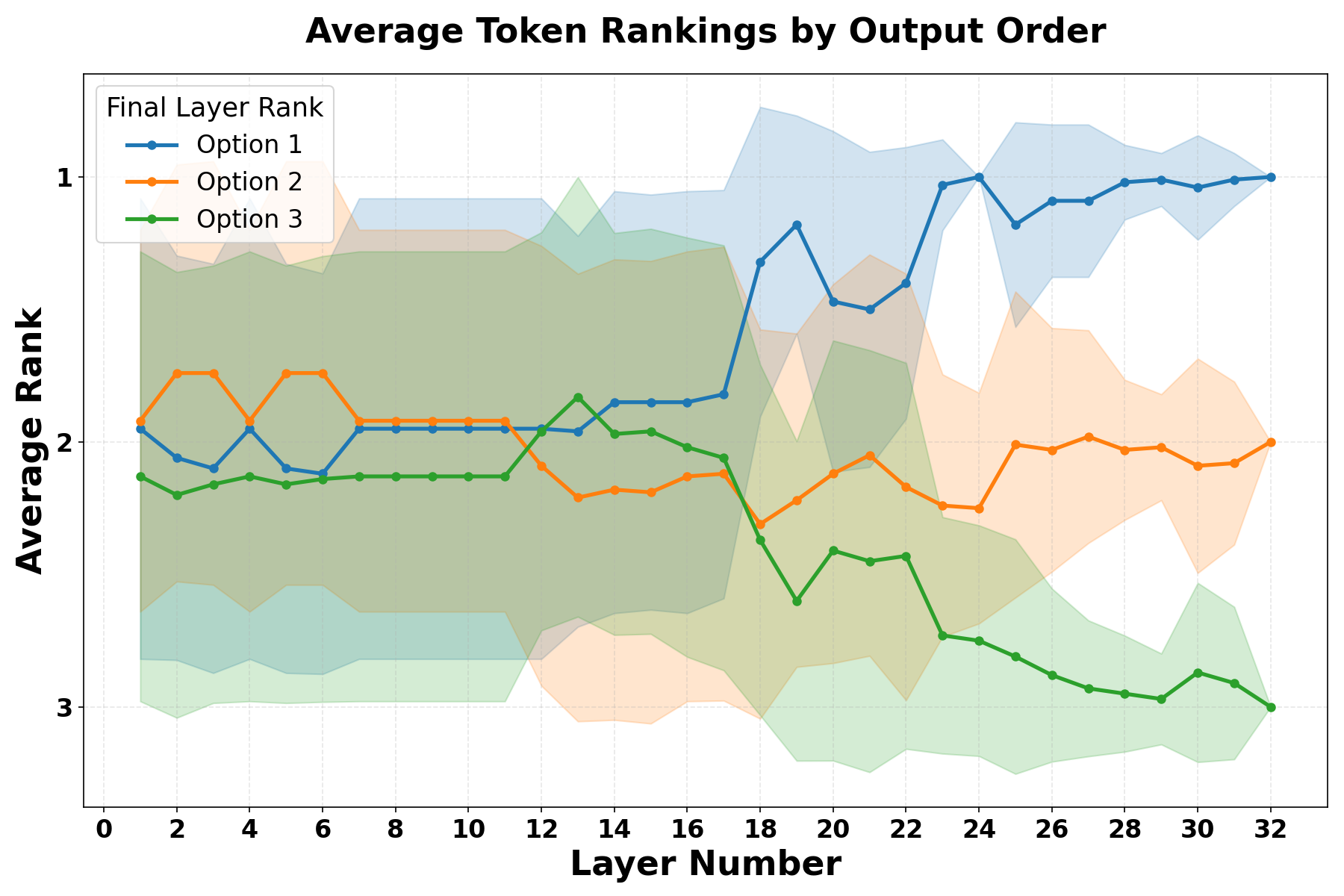}
\subcaption{Llama (3 Options)}
\end{subfigure}
\hfill
\begin{subfigure}{0.48\textwidth}
\centering
\includegraphics[width=\textwidth]{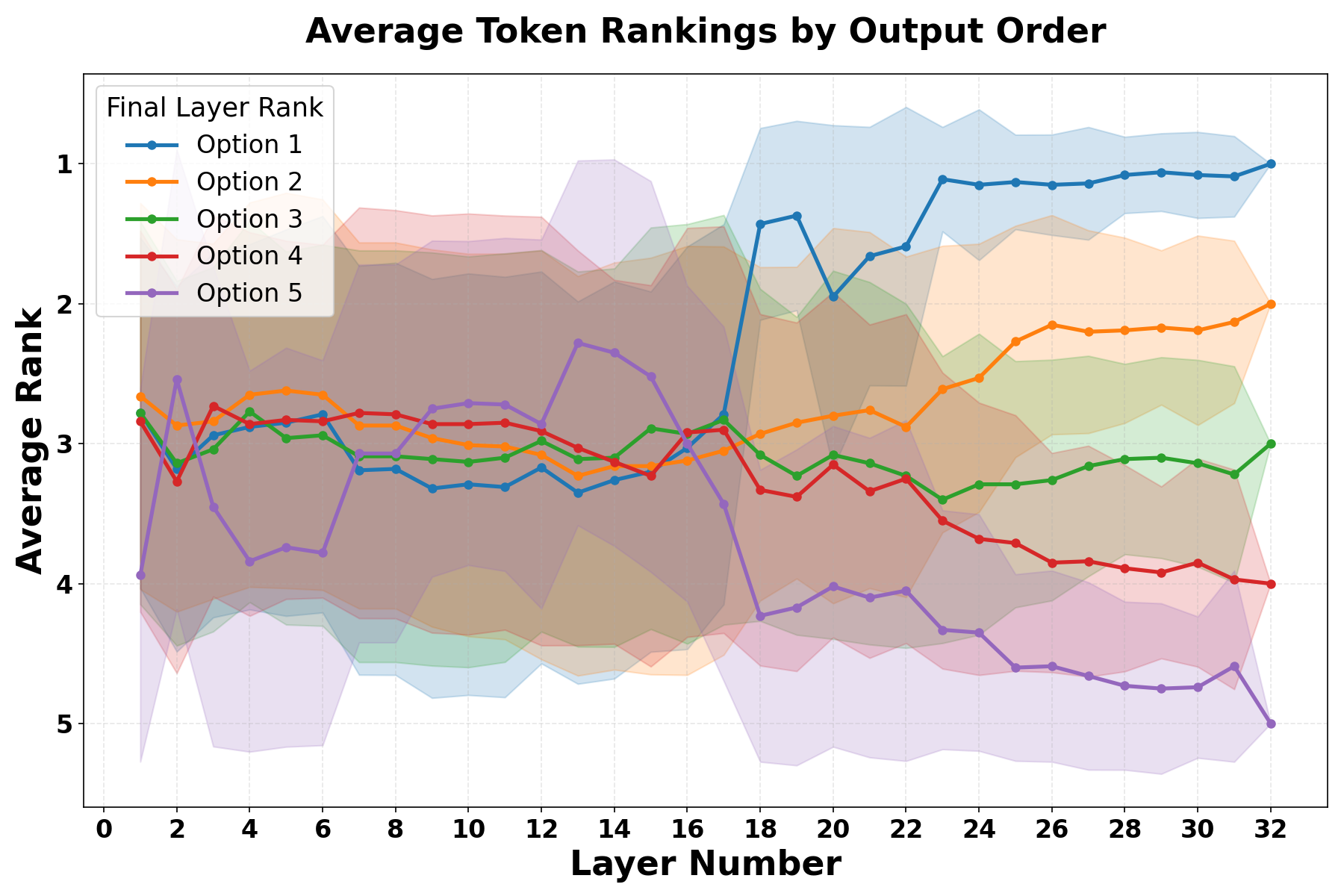}
\subcaption{Llama (5 Options)}
\end{subfigure}
\caption{Average token ranking plots for varying numbers in the multiple-choice options for Qwen and Llama models on the QASC dataset.}
\label{fig:number_of_options_aggregate}
\end{figure*}

\begin{figure*}[!h]
\centering
\begin{subfigure}{0.48\textwidth}
\centering
\includegraphics[width=\textwidth]{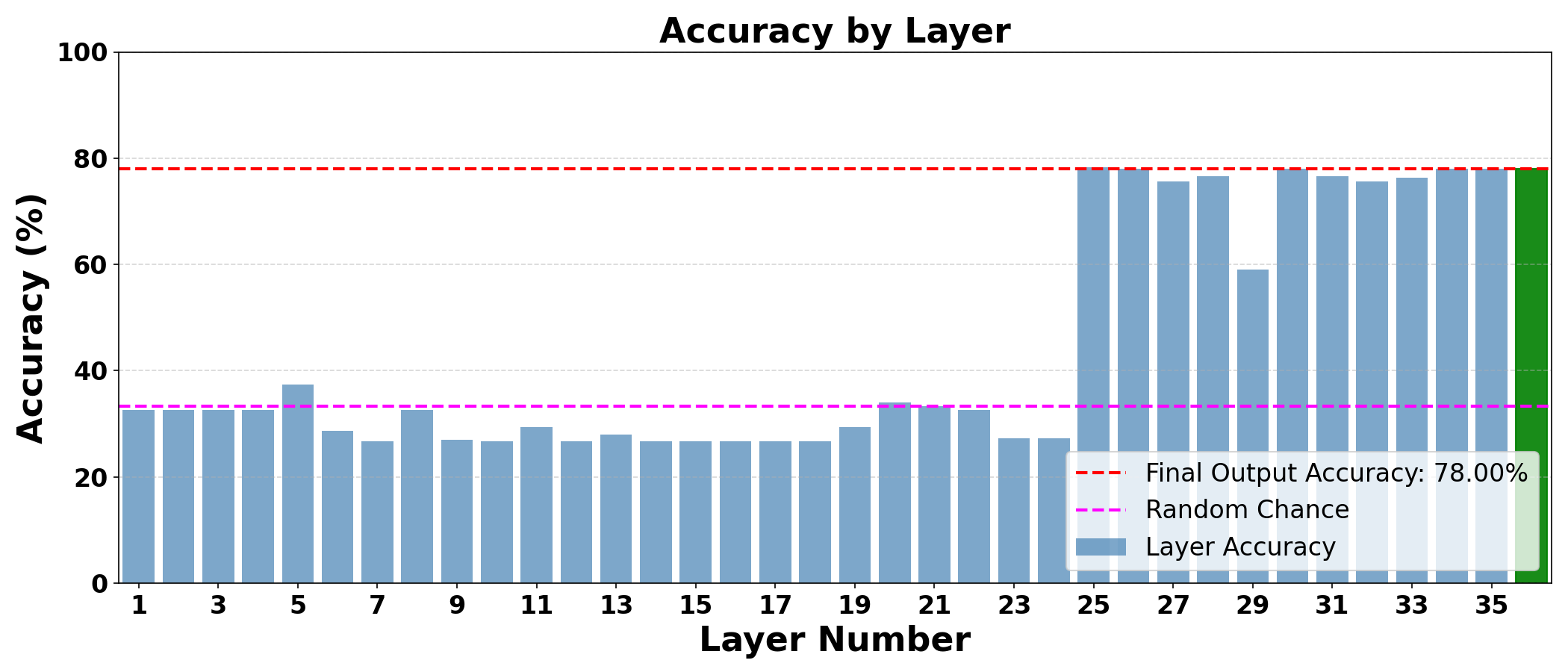}
\subcaption{Qwen (3 Options)}
\end{subfigure}
\hfill
\begin{subfigure}{0.48\textwidth}
\centering
\includegraphics[width=\textwidth]{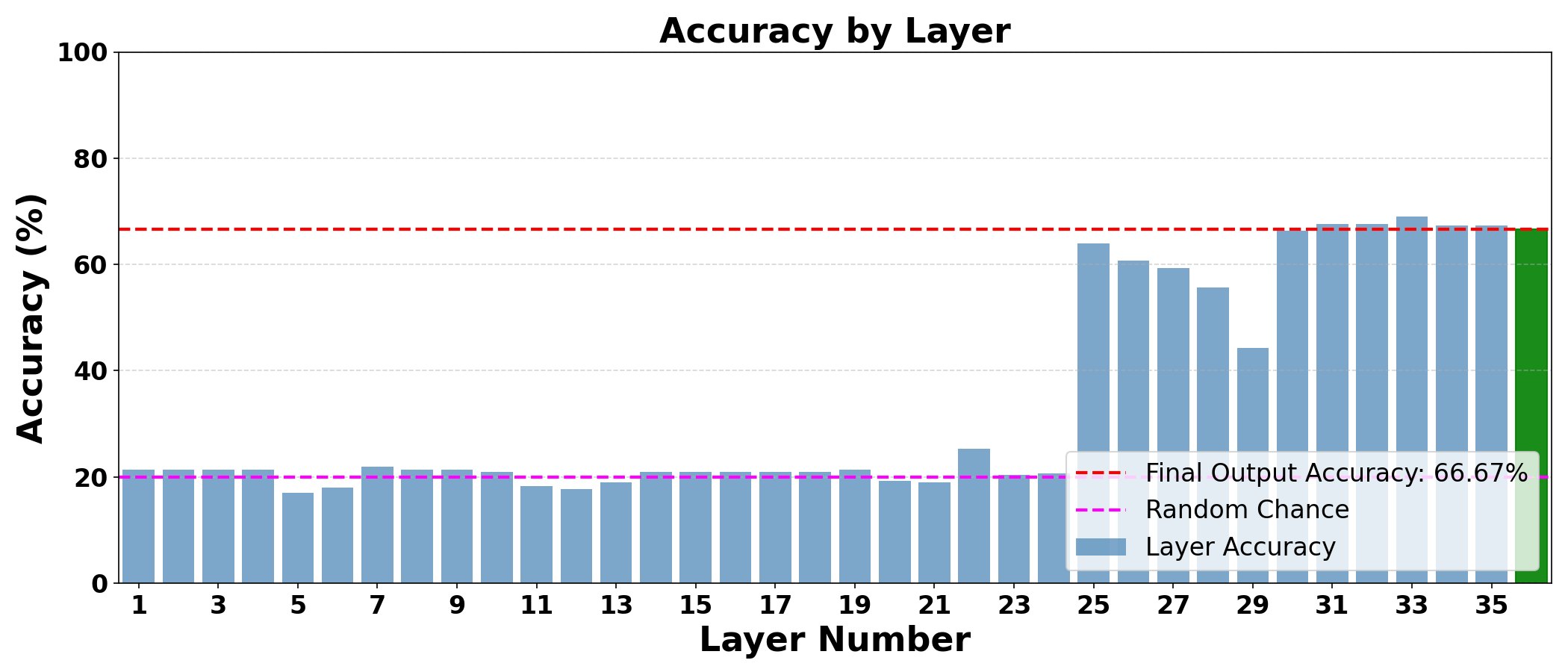}
\subcaption{Qwen (5 Options)}
\end{subfigure}
\\
\begin{subfigure}{0.48\textwidth}
\centering
\includegraphics[width=\textwidth]{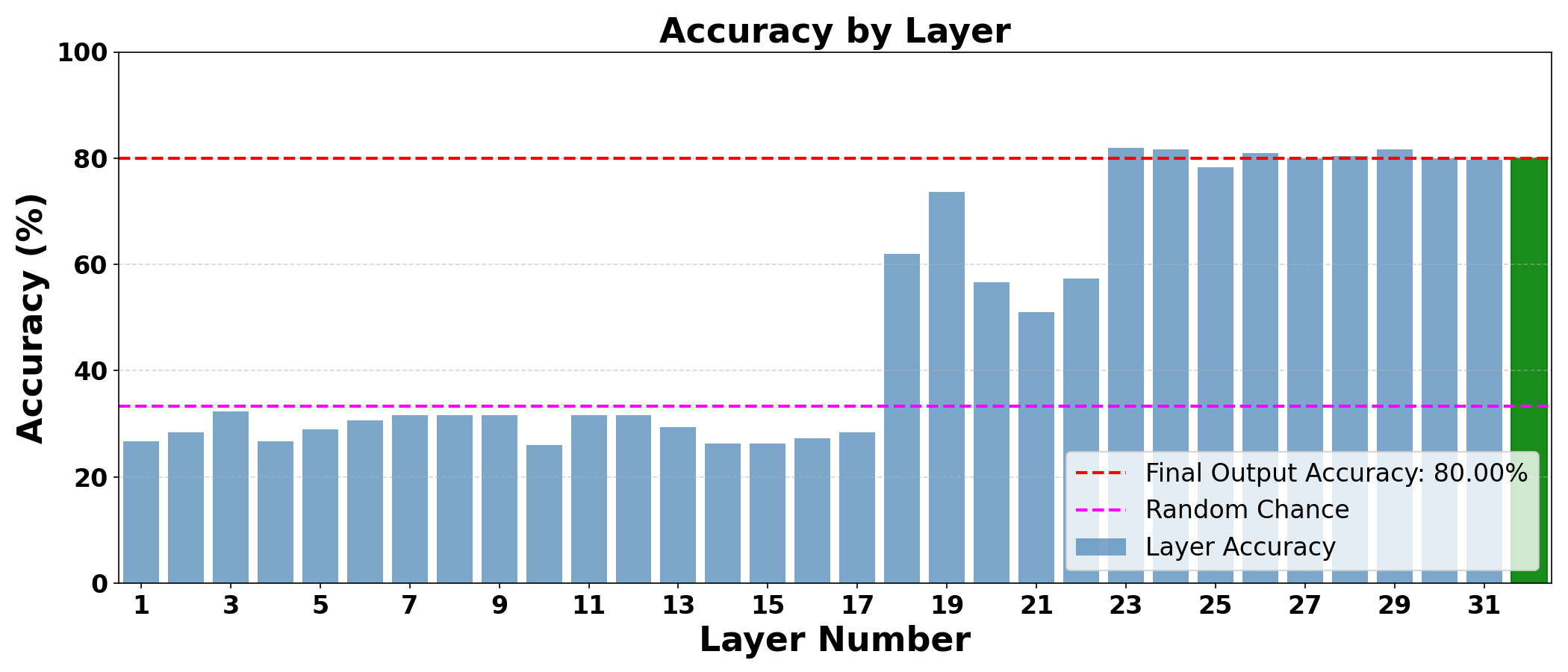}
\subcaption{Llama (3 Options)}
\end{subfigure}
\hfill
\begin{subfigure}{0.48\textwidth}
\centering
\includegraphics[width=\textwidth]{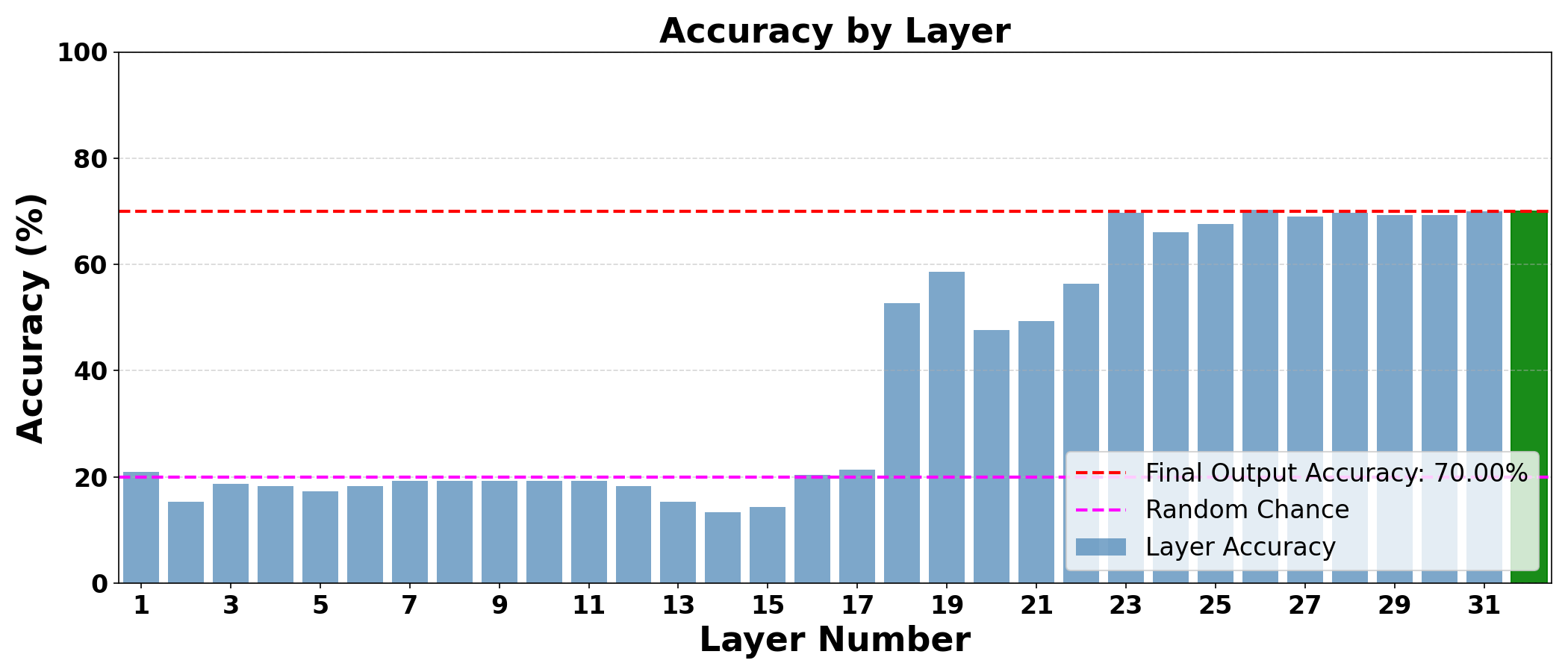}
\subcaption{Llama (5 Options)}
\end{subfigure}
\caption{Layer-wise accuracy comparison across different numbers of multiple-choice options.}
\label{fig:number_of_options_accuracy}
\end{figure*}

\begin{figure*}[h!]
\centering
\begin{subfigure}{0.48\textwidth}
\centering
\includegraphics[width=\textwidth]{images/plots_for_paper/qasc/qwen/aggregate_plots/Correct/aggregate_overall_by_output_order.png}
\subcaption{Qwen (Alphabets)}
\end{subfigure}
\hfill
\begin{subfigure}{0.48\textwidth}
\centering
\includegraphics[width=\textwidth]{images/plots_for_paper/qasc/llama/aggregate_plots/Correct/aggregate_overall_by_output_order.png}
\subcaption{Llama (Alphabets)}
\end{subfigure}
\\
\begin{subfigure}{0.48\textwidth}
\centering
\includegraphics[width=\textwidth]{images/other_option_types/qasc_arabic/qwen/aggregate_plots/Correct/aggregate_overall_by_output_order.png}
\subcaption{Qwen (Arabic)}
\end{subfigure}
\hfill
\begin{subfigure}{0.48\textwidth}
\centering
\includegraphics[width=\textwidth]{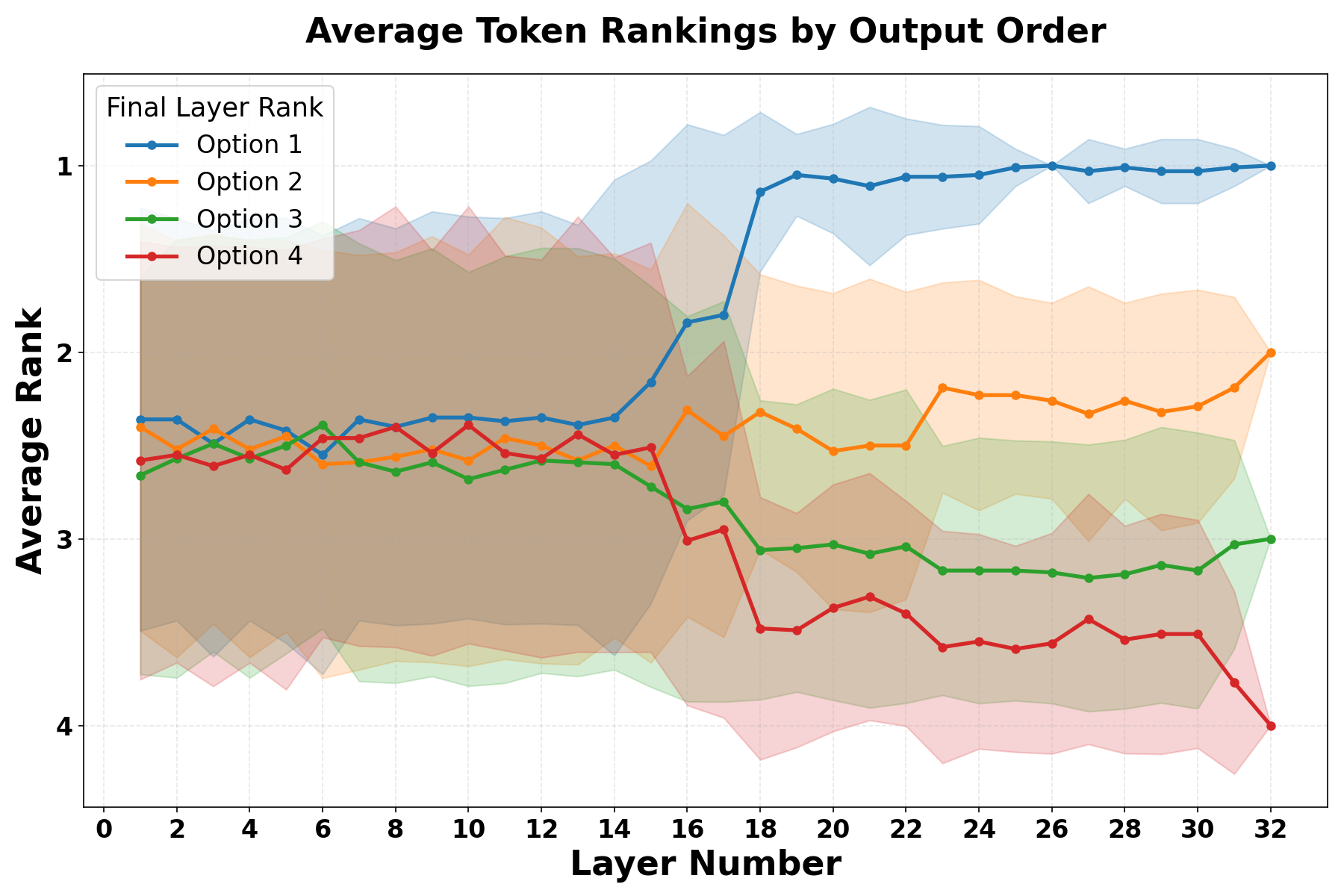}
\subcaption{Llama (Arabic)}
\end{subfigure}
\\
\begin{subfigure}{0.48\textwidth}
\centering
\includegraphics[width=\textwidth]{images/other_option_types/qasc_roman/qwen/aggregate_plots/Correct/aggregate_overall_by_output_order.png}
\subcaption{Qwen (Roman)}
\end{subfigure}
\hfill
\begin{subfigure}{0.48\textwidth}
\centering
\includegraphics[width=\textwidth]{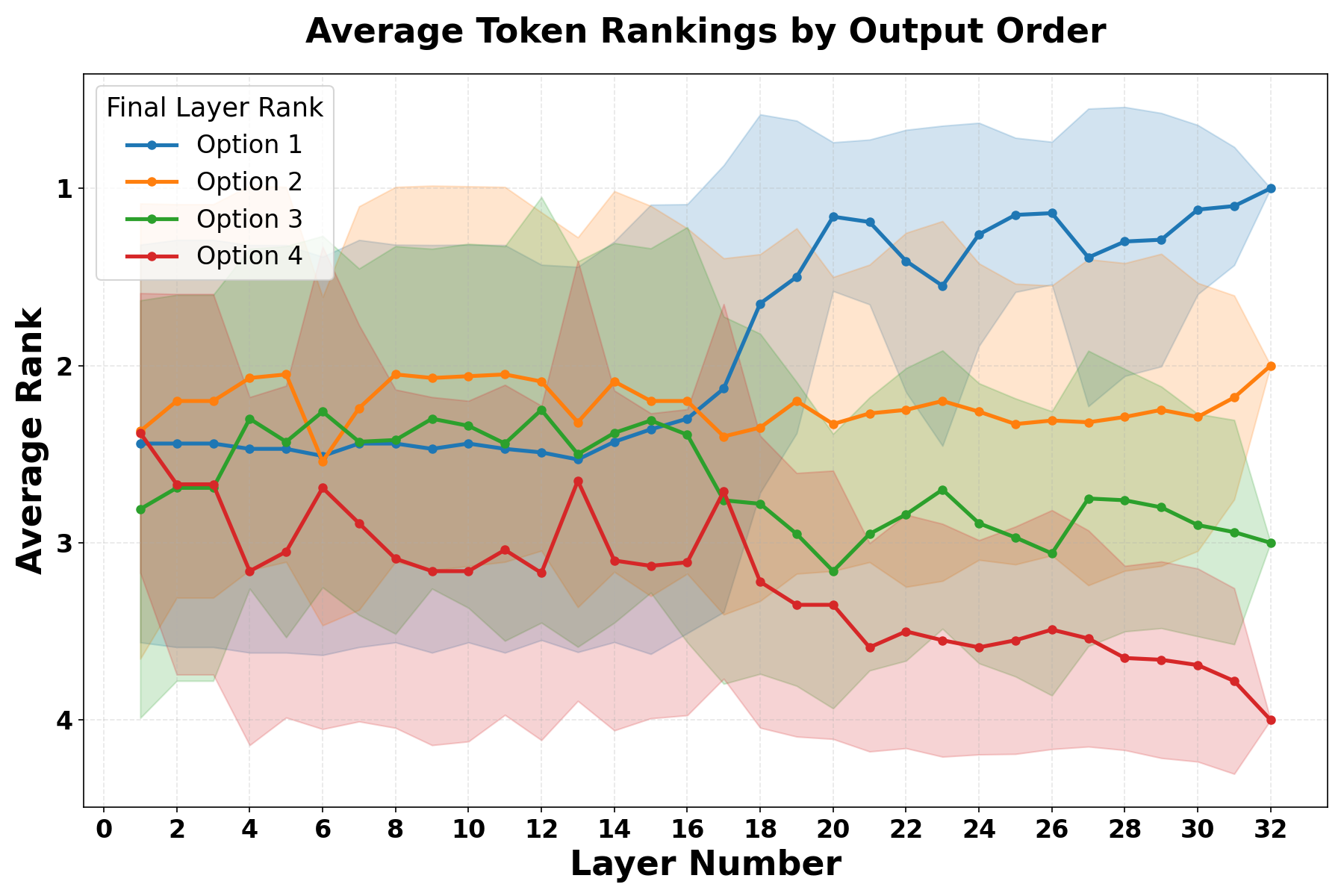}
\subcaption{Llama (Roman)}
\end{subfigure}
\caption{Impact of option labeling schemes on token rankings across layers. Each subfigure shows aggregate token ranking plots for either Qwen or Llama across three labeling schemes: Alphabets (A/B/C/D), Arabic numerals (1/2/3/4), and Roman numerals (i/ii/iii/iv), all evaluated on the QASC dataset. The plots reveal whether the location and prominence of the Hard Decision Layer (HDL) varies across different labeling schemes. Generally, the HDL manifests at a similar layer position regardless of labeling scheme, though its clarity differs: Alphabets and Arabic numerals produce a well-pronounced HDL, while Roman numerals show a less distinct pattern.}
\label{fig:option_labels_aggregate_all}
\end{figure*}

\begin{figure*}[!h]
\centering
\begin{subfigure}{0.48\textwidth}
\centering
\includegraphics[width=\textwidth]{images/plots_for_paper/qasc/qwen/accuracy_by_layer.png}
\subcaption{Qwen (Alphabets)}
\end{subfigure}
\hfill
\begin{subfigure}{0.48\textwidth}
\centering
\includegraphics[width=\textwidth]{images/plots_for_paper/qasc/llama/accuracy_by_layer.png}
\subcaption{Llama (Alphabets)}
\end{subfigure}
\\
\begin{subfigure}{0.48\textwidth}
\centering
\includegraphics[width=\textwidth]{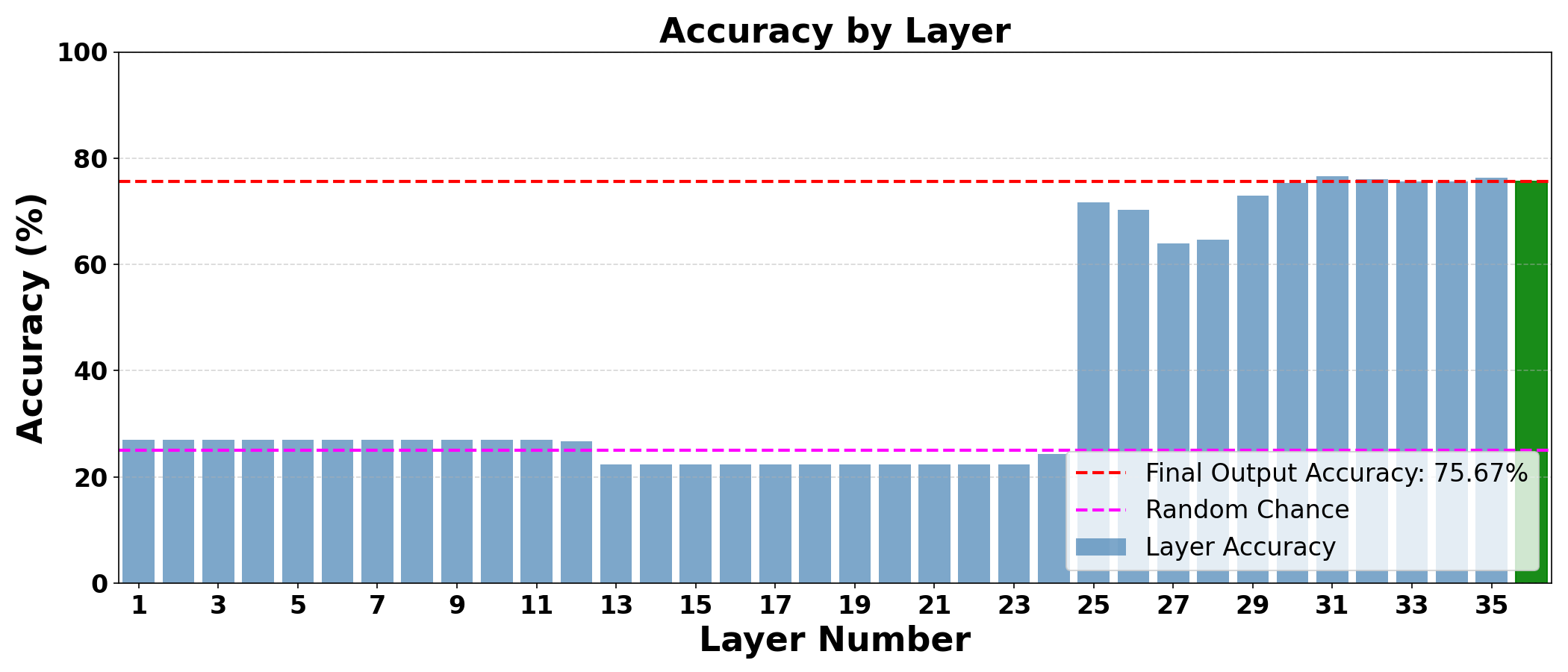}
\subcaption{Qwen (Arabic)}
\end{subfigure}
\hfill
\begin{subfigure}{0.48\textwidth}
\centering
\includegraphics[width=\textwidth]{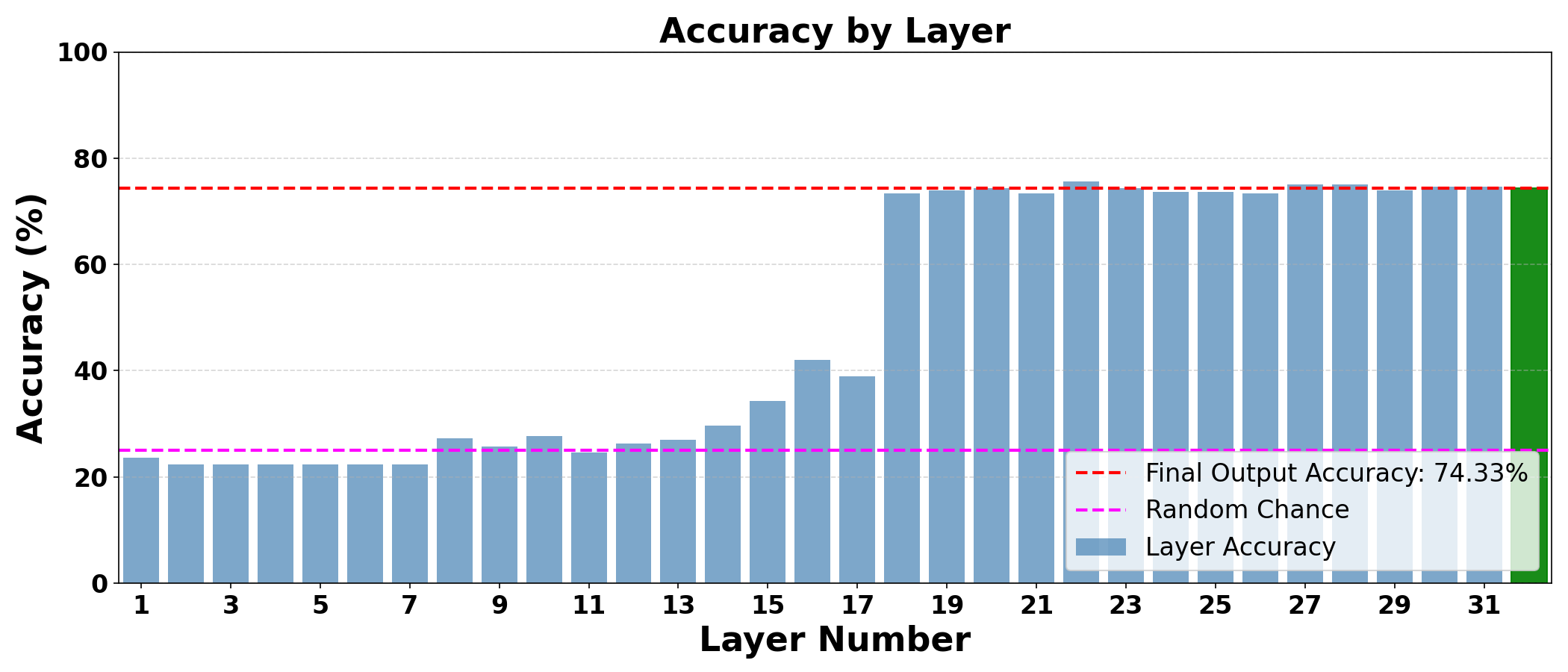}
\subcaption{Llama (Arabic)}
\end{subfigure}
\\
\begin{subfigure}{0.48\textwidth}
\centering
\includegraphics[width=\textwidth]{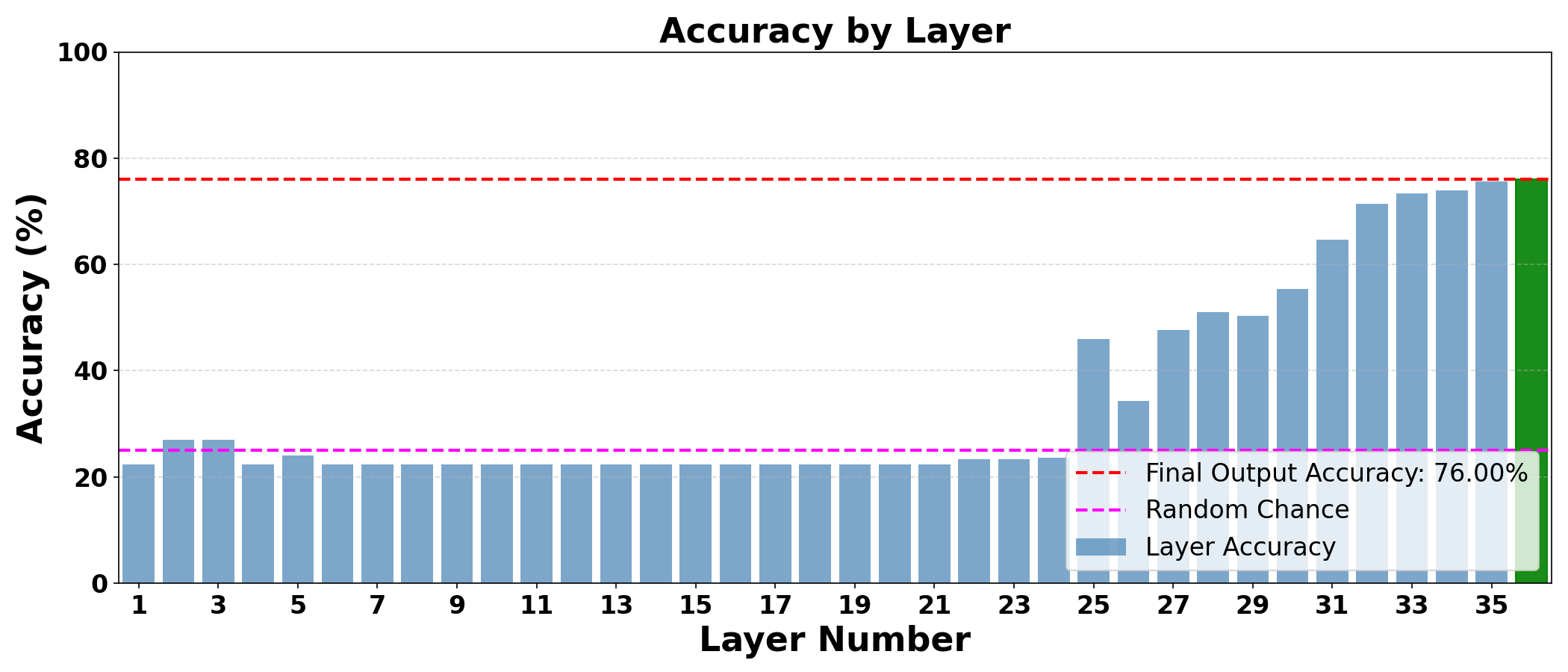}
\subcaption{Qwen (Roman)}
\end{subfigure}
\hfill
\begin{subfigure}{0.48\textwidth}
\centering
\includegraphics[width=\textwidth]{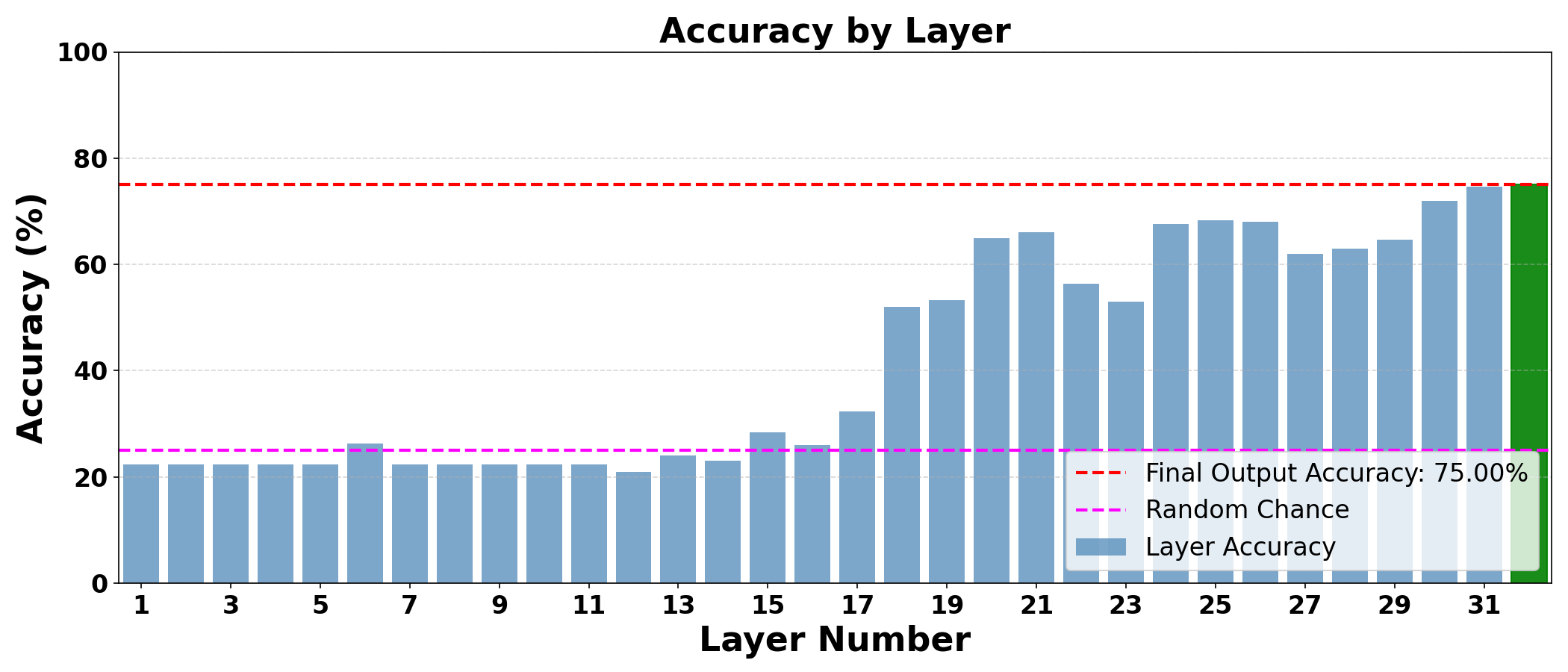}
\subcaption{Llama (Roman)}
\end{subfigure}
\caption{Layer-wise accuracy comparison across different option label types.}
\label{fig:option_labels_accuracy}
\end{figure*}

Figures~\ref{fig:number_of_options_aggregate} and~\ref{fig:number_of_options_accuracy} show the average token rankings and accuracy through the layers for the Qwen and Llama models when evaluated on the QASC dataset with varying number of answer choices. Figures~\ref{fig:option_labels_aggregate_all} and~\ref{fig:option_labels_accuracy} along with Table~\ref{tab:hdl_label_variants} show the results of Qwen and Llama models when the option labels were varied.

\begin{table*}[!h]
\centering
\small
\begin{tabular}{cccccc}
\toprule
\textbf{Model} & \textbf{Label Type} & \makecell{\textbf{HDL Layer} \\ \textbf{(Predicted)}} & \makecell{\textbf{Accuracy} \\ \textbf{(Pre-HDL)}} & \makecell{\textbf{Accuracy} \\ \textbf{(Post-HDL)}} & \makecell{\textbf{Accuracy} \\ \textbf{(Final)}} \\
\midrule
Qwen & alpha & 25 & 0.31 & 0.70 (+0.39) & 0.72 (+0.02) \\
Qwen & arabic & 25 & 0.31 & 0.73 (+0.42) & 0.75 (+0.02) \\
Qwen & roman & 25 & 0.27 & 0.57 (+0.30) & 0.77 (+0.20) \\
\hline
Llama & alpha & 18 & 0.28 & 0.59 (+0.31) & 0.74 (+0.15) \\
Llama & arabic & 18 & 0.43 & 0.68 (+0.25) & 0.71 (+0.03) \\
Llama & roman & 18 & 0.42 & 0.56 (+0.14) & 0.73 (+0.17) \\
\bottomrule
\end{tabular}
\caption{Descriptive HDL summary across QASC option-label variants. All variants present 4 options on the same QASC questions; only the Label Type differs (alpha = A/B/C/D, arabic = 1/2/3/4, roman = i/ii/iii/iv). Each row shows the HDL Layer (Predicted) and accuracies at the layer just before HDL (Accuracy (Pre-HDL)), at the HDL layer (Accuracy (Post-HDL)), and at the final layer (Accuracy (Final)) for one (model, label type) combination; parenthetical values on the accuracy columns are absolute increases over the previous column.}
\label{tab:hdl_label_variants}
\end{table*}

\section{Plots for Open-Ended Generation}\label{sec:plots-open-ended}
\begin{figure*}[!h]
\centering
\begin{subfigure}{0.48\textwidth}
\centering
\includegraphics[width=\textwidth]{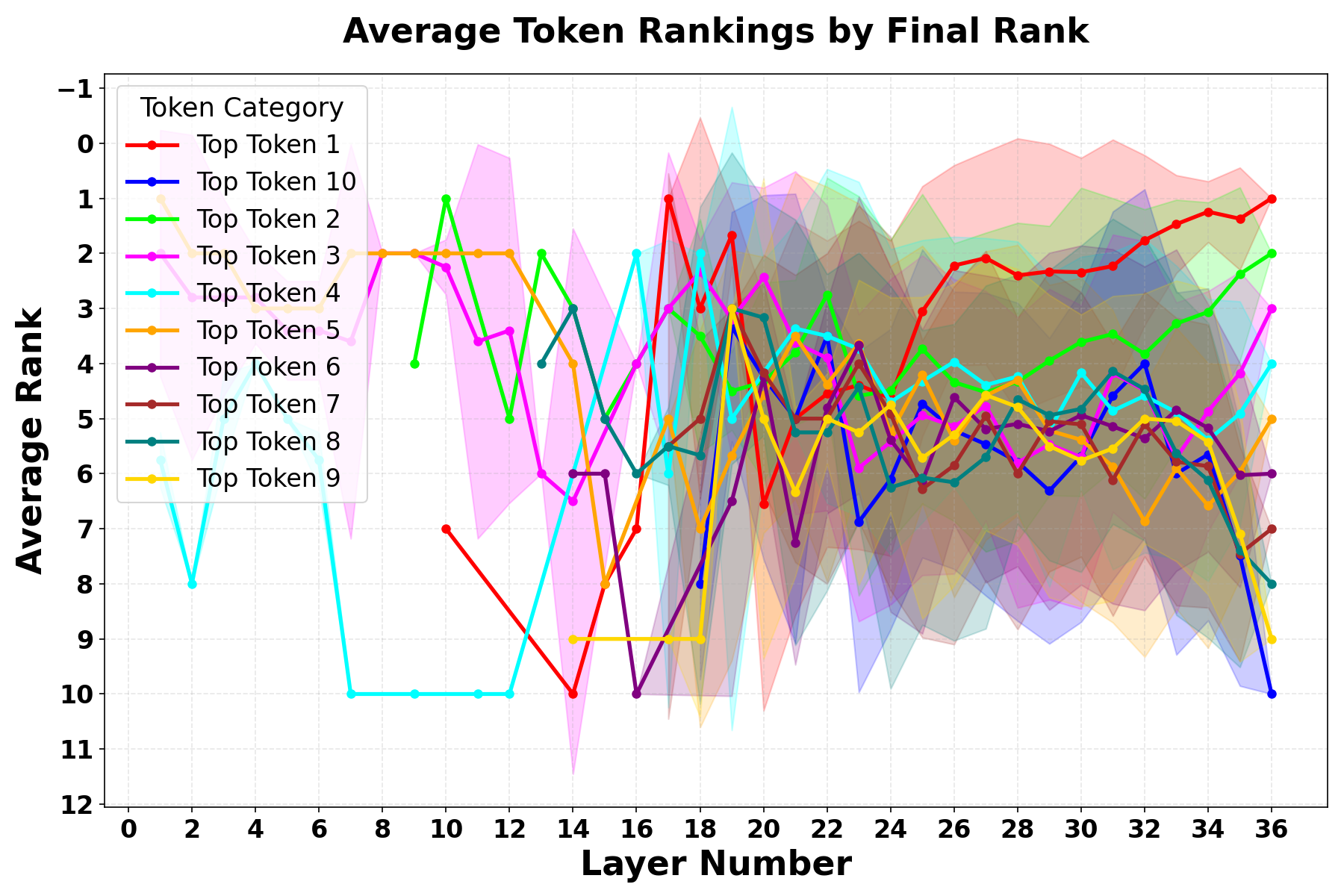}
\subcaption{$n=10$}
\end{subfigure}
\hfill
\begin{subfigure}{0.48\textwidth}
\centering
\includegraphics[width=\textwidth]{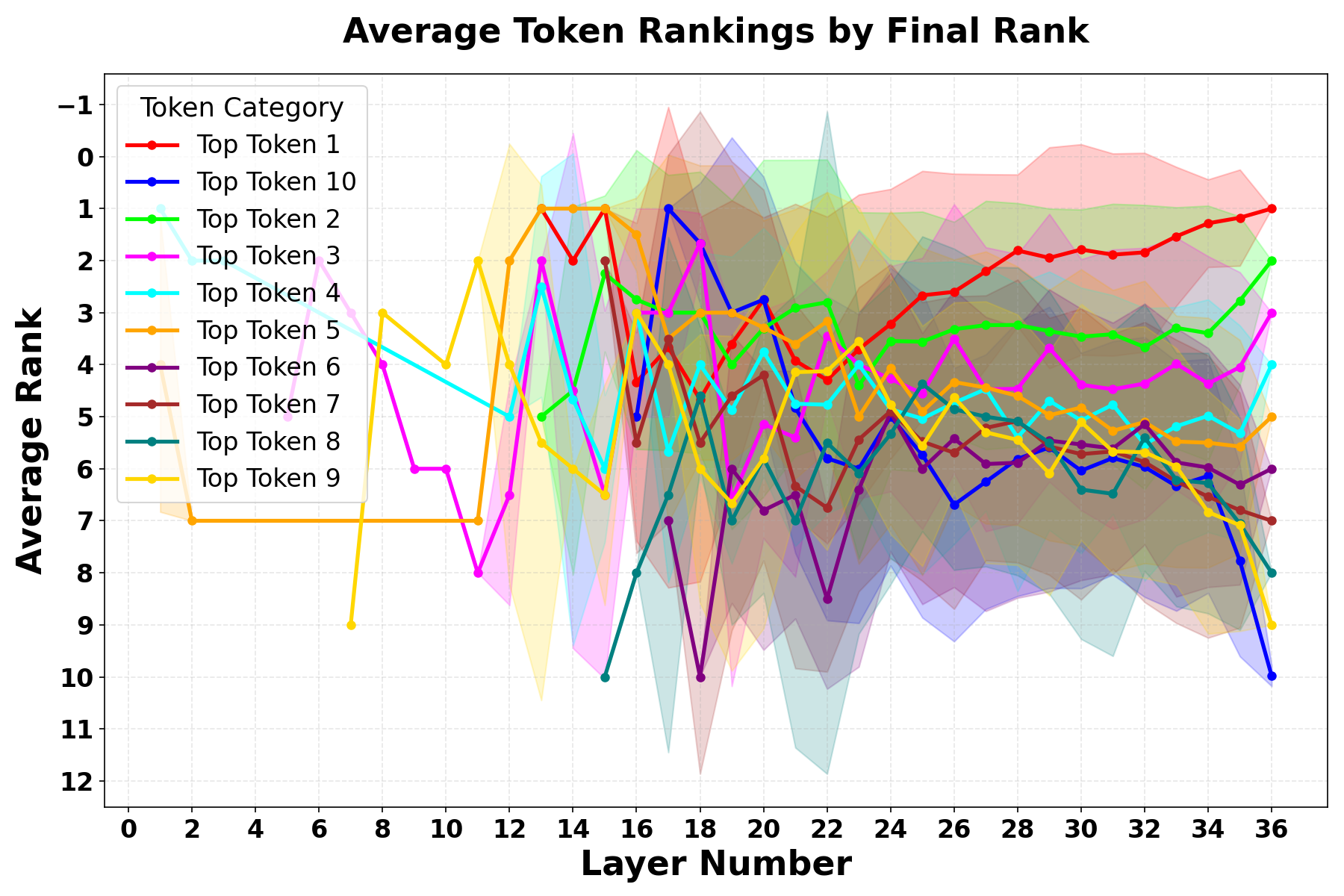}
\subcaption{$n=50$}
\end{subfigure}
\vspace{0.5cm}

\begin{subfigure}{0.48\textwidth}
\centering
\includegraphics[width=\textwidth]{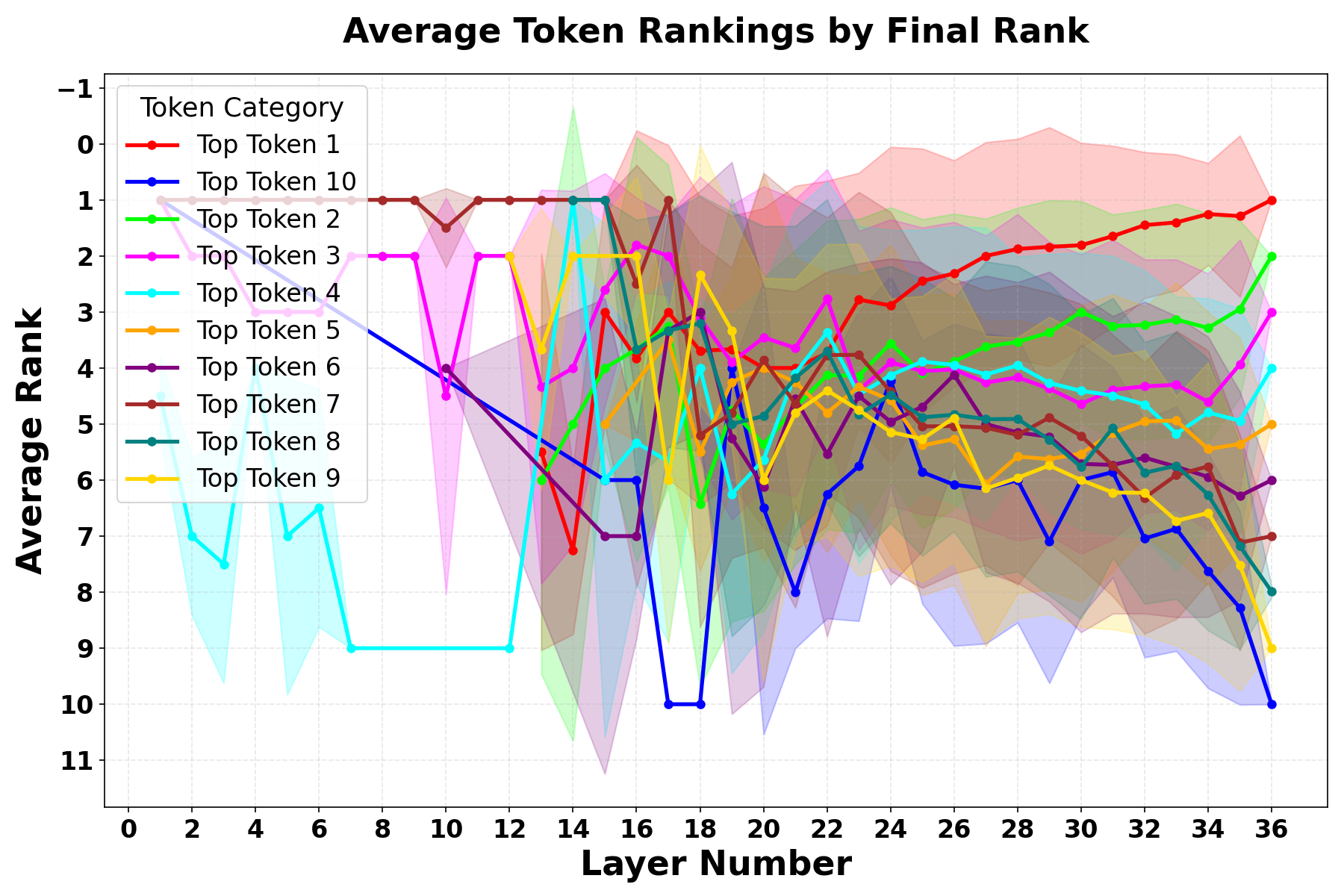}
\subcaption{$n=100$}
\end{subfigure}
\hfill
\begin{subfigure}{0.48\textwidth}
\centering
\includegraphics[width=\textwidth]{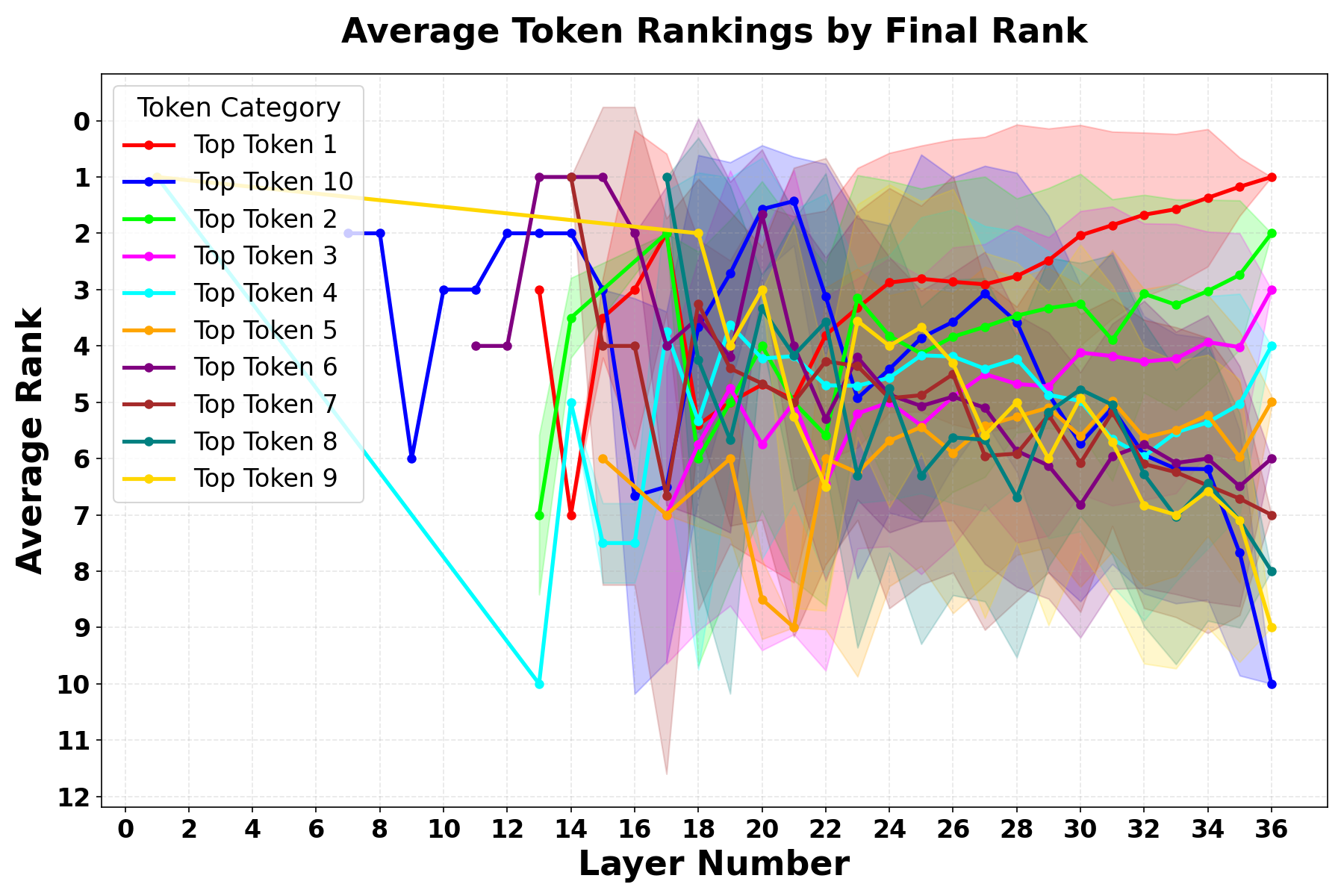}
\subcaption{$n=200$}
\end{subfigure}

\caption{Aggregate plots of $n$th token generation in the open-ended GSM8K evaluation for Qwen.}
\label{fig:openended_generation}
\end{figure*}

Figure~\ref{fig:openended_generation} shows the average token rankings through the layers for the Qwen model on the open-ended generation task evaluated on different token positions $n = {10, 50, 100, 200}$




\end{document}